\documentclass[pdflatex,sn-mathphys-num]{sn-jnl}

\usepackage{graphicx}%
\usepackage{multirow}%
\usepackage{amsmath,amssymb,amsfonts}%
\usepackage{amsthm}%
\usepackage{mathrsfs}%
\usepackage[title]{appendix}%
\usepackage{textcomp}%
\usepackage{manyfoot}%
\usepackage{booktabs}%
\usepackage{algorithm}%
\usepackage{algorithmicx}%
\usepackage{algpseudocode}%
\usepackage{listings}%
\usepackage{tabularx}
\usepackage[table]{xcolor}
\usepackage{tikz}

\newcommand{\greencheck}{{\color{green}\checkmark}}
\newcommand{\redcross}{{\color{red}\times}}
\newcommand{\greenup}{{\color{green}\uparrow}}
\newcommand{\reddown}{{\color{red}\downarrow}}

\definecolor{lightblue}{RGB}{173, 216, 230}
\definecolor{lavender}{RGB}{230, 230, 250}

\theoremstyle{thmstyleone}%
\theoremstyle{thmstyletwo}%

\theoremstyle{thmstylethree}%

\begin{document}

\title[Humanlike cognition in LLMs]{Humanlike Cognitive Patterns as Emergent Phenomena in Large Language Models}


\author[1]{\fnm{Zhisheng} \sur{Tang}}\email{zhisheng@isi.edu}

\author*[1]{\fnm{Mayank} \sur{Kejriwal}}\email{kejriwal@isi.edu}

\affil*[1]{\orgdiv{Information Sciences Institute}, \orgname{USC Viterbi School of Engineering}, \orgaddress{\street{4676 Admiralty Way 1001}, \city{Marina Del Rey}, \postcode{90292}, \state{California}, \country{USA}}}


\abstract{
Research on emergent patterns in Large Language Models (LLMs) has gained significant traction in both psychology and artificial intelligence, motivating the need for a comprehensive review that offers a synthesis of this complex landscape. In this article, we systematically review LLMs’ capabilities across three important cognitive domains (decision-making biases, reasoning, and creativity), using empirical studies drawing on established psychological tests and comparing LLMs' performance to human benchmarks. On decision-making, our synthesis reveals that while LLMs demonstrate several human-like biases,  some biases observed in humans are absent, indicating cognitive patterns that only partially align with human decision-making. On reasoning, advanced LLMs like GPT-4 exhibit deliberative reasoning akin to human System-2 thinking, while smaller models fall short of human-level performance. A distinct dichotomy emerges in creativity: while LLMs excel in language-based creative tasks, such as storytelling, they struggle with divergent thinking tasks that require real-world context. Nonetheless, studies suggest that LLMs hold considerable potential as collaborators, augmenting creativity in human-machine problem-solving settings. Discussing key limitations, we also offer guidance for future research in areas such as memory, attention, and open-source model development.
}




\maketitle

\section{Introduction}

Recent advancements in generative artificial intelligence (AI) and large language models (LLMs) \cite{brown2020language,achiam2023gpt} have sparked new interest in how AI might simulate or even influence human cognition and decision-making. Although the original purpose of language models was text generation and addressing longstanding problems in natural language processing (NLP) like machine translation and information extraction \cite{vaswani2017attention, popel2020transforming, bahdanau2014neural, huang2015bidirectional, lin2016neural}, a recent suite of papers have argued that they are now exhibiting, or at the very least, mimicking, complex reasoning abilities at human levels of performance \cite{park2023generative, strachan2024testing, bubeck2023sparks}. Beyond reasoning and decision-making, LLMs released since the first edition of ChatGPT in early 2023 have also been argued to mimic abilities like \textit{creativity} (evidenced through tasks like poetry and lyrical generation \cite{sarrion2023using}), although the originality of their creative outputs is a matter of debate.

These debates notwithstanding, the \textit{emergence} of novel behaviors and properties in LLMs as an empirical phenomenon cannot be ignored \cite{wei2022emergent}. Researchers are divided on the causes and implications of such behaviors, with some arguing that they are largely a mirage \cite{schaeffer2024emergent} and others arguing that true emergence is occurring \cite{wei2022emergent}. By \textit{emergence} here, we simply mean that the model was not explicitly trained to mimic or learn such behaviors, either through the underlying neural network's objective function or through the data itself. \citet{wei2022emergent} succinctly describe emergence in the context of LLMs as an ability that is not present in smaller models but is present in larger models.

While emergence is much more complex in LLMs, it has some precedent in NLP research since 2010 that has increasingly relied upon deep neural networks. For example, when the word2vec model first became popular more than a decade ago, the authors of the original paper noted how the word vector representations yielded by the skip-gram neural network model (when it was fed reasonably large corpora in an unsupervised fashion) obeyed the now-classic analogy $\vec{king}-\vec{man}+\vec{woman}=\vec{queen}$ \cite{mikolov2013efficient}. Similarly, shortly following the release of Bidirectional Encoder Representations from Transformers (BERT) \cite{devlin2018bert}, one of the first transformer-based language models, a line of work colloquially referred to as BERTology \cite{rogers2021primer} rapidly emerged in less than five years, showcasing empirical phenomenon of a largely emergent nature. 
For example, BERT’s neural layers were found to exhibit hierarchical representations of language (despite not being explicitly trained to do so), with earlier layers containing information about linear word order, middle layers carrying syntactic information, final layers holding task-specific knowledge, and semantics spreading across all layers.

Given these expanded and unexpected capabilities, an intriguing question arises: do LLMs, which are trained on vast amounts of human-generated text, also exhibit cognitive patterns that are typically shown in humans? Specifically, do they exhibit the cognitive heuristics and biases that characterize human decision-making? Do they share the same kinds of reasoning patterns and levels of reasoning capabilities as humans? Can they innovate in ways that resemble human creativity? To answer these questions, it is helpful to first understand what these cognitive patterns are, and the relationship between them. 

Cognitive scientists and psychologists have long studied human decision-making biases, such as hindsight bias, overweighting, and belief bias. Although often reducing cognitive effort, they can also lead to suboptimal decisions \cite{shah2008heuristics}. However, these biases are just one element in the interconnected cognitive processes that underlie human cognition. Decision-making itself is the process of selecting from various alternatives by weighing their respective benefits and drawbacks \cite{kahneman1979prospect}. In this context, reasoning is considered the ability to logically evaluate the pros and cons of different alternatives \cite{baron2023thinking}. Simultaneously, creativity can be defined as the ability to generate novel alternatives and enable decision-making by introducing innovative solutions \cite{amabile2018creativity}. Together, reasoning helps evaluate alternatives, while creativity supports generating new alternatives, making them both essential for effective decision-making. Investigating LLMs' emergent behavior in these areas can offer a comprehensive view of their potential to replicate human-like cognitive processes.

\begin{figure}[h]
    \centering
    \includegraphics[width=\textwidth]{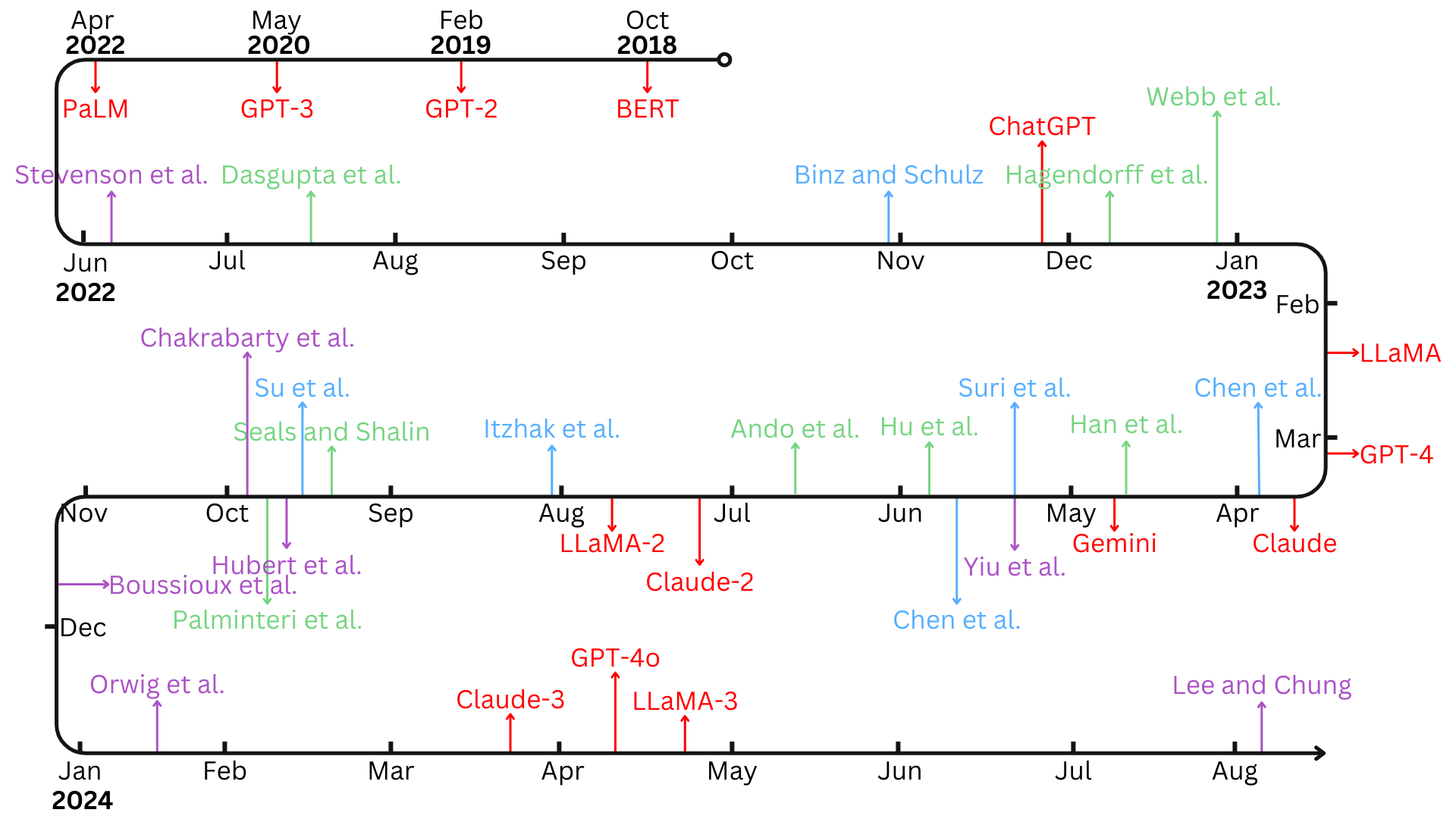}
    \caption{Timeline of various studies reviewed, along with release time of major Large Language Models. \textcolor{red}{Red}, \textcolor[rgb]{0.34, 0.68, 1.0}{Blue}, \textcolor[rgb]{0.46, 0.83, 0.51}{Green}, and \textcolor[rgb]{0.67, 0.32, 0.77}{Purple} entries indicate LLM releases, decision-making related studies, reasoning related studies, and creativity related studies, respectively.}
    \label{fig:timeline}
\end{figure}

As shown in Figure \ref{fig:timeline}, there is a line of studies that use established psychological and cognitive tools to examine how LLMs replicate or diverge from human patterns in the three cognitive processes of \textit{decision-making, reasoning,} and \textit{creativity}. In LLMs, these processes are better denoted as \textit{patterns}, because the actual mechanisms through which LLMs are able to replicate such abilities are obviously different from the way in which humans do so, at least neurally. We include studies that investigate the extent to which LLMs demonstrate human-like decision-making biases, their capability to reason under varied conditions, and their ability to innovate. We do not include studies that are based completely on computational tests that have not undergone rigorous human benchmarking of some sort. Rather, we focus solely on studies that use established psychological experiments or their variants, and for which human baseline performance is available. This allows us to make clear comparisons between LLMs and humans and ensures that any critiques or insights offered are grounded in robust experimental results with a human performance reference.

Overall, this review attempts a comprehensive examination of LLMs' performance across the three cognitive processes that seem to underlie emergence that have relevance to psychology research. For each process, we begin by briefly introducing the process and its fundamental principles. We then review individual studies that have attempted to empirically study or measure that process in at least one LLM, discussing their methodologies and findings. Finally, we conclude each section by synthesizing the insights, comparing LLMs’ behaviors with human performance, and reflecting on the implications of the findings. By systematically comparing LLMs’ behaviors to established human benchmarks, we aim to synthesize the growing body of literature on how these models simulate human cognitive patterns. We close by providing guidance for future research that is needed in this area to draw firmer conclusions.

\section{Decision-Making Cognitive Patterns}

\begin{table}[!htp]\centering\caption{Decision-making related studies, with specific type, dataset/tasks used, LLMs involved in the study, associated findings, and corresponding citation. $\greencheck$ denotes biases/heuristics found, while $\redcross$ denotes absence.}\label{tab:dm}
\begin{tabularx}{\columnwidth}{Xp{2.5cm}Xp{5cm}X}
\hline
Type & Dataset & LLMs & Findings & Citation \\
\hline
Bias & 12 vignette-based experiments and a set of multiple choice questions inspired by \cite{kahneman1972subjective} & GPT-3 & Framing effect\greencheck, certainty effect\greencheck, overweighting bias\greencheck, reﬂection effect$\redcross$, isolation effect$\redcross$, magnitude perception$\redcross$ & \citet{binz2023a} \\
Bias &Curated dataset &GPT-3, GPT-3.5, GPT-4 etc. & Decoy effect\greencheck, certainty effect\greencheck, belief bias\greencheck &\citet{itzhak2023instructed} \\
Bias &Curated decision-making problems and derivatives that are rewritten in operation management context &GPT-3.5, GPT-4 & Risk aversion\greencheck, preference for certainty in subjective tasks\greencheck, heuristic reasoning in objective tasks\greencheck &\citet{chen2023manager} \\
Bias &Newsvendor problem &GPT-4 & Demand-chasing\greencheck, risk-aversion\greencheck, loss-aversion\greencheck, waste aversion$\redcross$, stockout aversion$\redcross$, underestimated opportunity costs$\redcross$, minimized ex-post inventory errors$\redcross$ &\citet{su2023can} \\
Rationality &Budgetary experiments &GPT-3.5 &The model outperformed humans but showed performance drops with non-standard price presentation&\citet{chen2023emergence} \\
Heuristics and bias &Curated dataset &GPT-3.5 &Anchoring Heuristic\greencheck, representativeness and availability heuristic\greencheck, framing effect\greencheck, the endowment effect\greencheck &\citet{suri2023large} \\
\hline
\end{tabularx}
\end{table}

\subsection{Heuristics and biases}
While there is much literature studying different aspects of human decision-making, such as: cognitive heuristics and biases \cite{tversky1974judgment}, dual-process theory \cite{jonathan2017dual}, social influence and group decision-making \cite{cialdini2004social}, and neuroeconomics and computational models of decision-making \cite{glimcher2004neuroeconomics}, current research on LLMs' decision-making behavior has mainly focused on cognitive heuristics and biases. To understand the relevance of these studies, it is helpful to first consider the foundational role of heuristics and biases in human decision-making.

Mental shortcuts, or heuristics, reduce the cognitive load of intensive processes \cite{shah2008heuristics}. Notable heuristics include the \textit{availability} heuristic, representativeness heuristic, and the \textit{anchoring} heuristic. The availability heuristic is our inclination to rely on information that comes to our mind quickly and easily when evaluating decisions \cite{tversky1973availability}. For example, people might judge the probability of a rare event, such as a plane crash, to be higher than it actually is if they have recently seen such an event on the news. Although the availability heuristic helps make decisions quicker, it also leads to an overestimation of unlikely but memorable events. 

The representativeness heuristic involves assessing probabilities based on the similarity between a sample and a larger population \cite{white1984representativeness}. One famous example is when people stereotypically deemed a young woman, described as deeply concerned with social justice, intelligent, and outspoken, to be more likely to be a bank teller and also active in the feminist movement, than to just be a bank teller alone (which would be inconsistent with rational rules of probability). The heuristic simplifies decision-making by relying on stereotypes but often misleads people to ignore general statistics. The anchoring heuristic is yet another commonly known mental shortcut. It occurs when the initial exposure to a reference point influences subsequent decisions \cite{furnham2011literature}. It is explored extensively in the science of framing sales as discounts. For example, if the original price of a commodity serves as an anchor, the discounted price suddenly becomes attractive, even though it may (still) be unreasonably high.

While these heuristics help individuals make decisions more effortlessly, they may also increase the odds of error, especially in situations of a slightly irregular nature \cite{gigerenzer2011heuristic}. In part, they cause a tendency to make simpler decisions and also exhibit biases. For instance, the hindsight bias can mislead people to see past events as more predictable after they have occurred. Similarly, overweighting bias causes people to assign disproportionate importance to certain pieces of information, often ones that are more recent or vivid, regardless of their actual relevance. Additionally, belief bias reflects a tendency to accept or reject conclusions based on alignment with pre-existing beliefs rather than logical validity.


\subsection{Evaluating cognitive heuristics and biases in LLMs} 

Table \ref{tab:dm} presents a comprehensive overview of heuristics and biases exhibited by different LLMs according to a range of studies. As an inspirational work, \citet{binz2023a} evaluated GPT-3’s decision-making capabilities using some classical cognitive psychology tasks, including vignette-based experiments (the Linda problem \cite{tversky1983extensional}, the cab problem \cite{tversky2015causal}), a recent task-based decision-making benchmark \cite{peterson2021using}, and a set of multiple choice questions design to elicit biases in human decision-making by \citet{kahneman1972subjective}. Through the vignette-based experiments, the authors discovered that, unlike people, GPT-3 did \textit{not} fall for the common fallacy but instead provided approximately correct answers. However, GPT-3’s performance declined drastically with minor changes to these vignettes, suggesting possible overfitting from pre-training. For the task-based investigation, even though GPT-3 was able to solve these problems above the chance level, it did not reach human-level performance. The work also found that GPT-3 showed three of six tested biases, including the framing effect, certainty effect, and overweighting bias.

Considering the mainstream approach for building consumer-facing LLMs, \citet{itzhak2023instructed} explored the impact of instruction-tuning and reinforcement learning from human feedback in LLMs on three cognitive biases: the decoy effect \cite{huber1982adding}, the certainty effect \cite{kahneman1979prospect}, and the belief bias \cite{evans1983conflict}. They found that these LLMs exhibit biases that align with \textit{human biases theory} \cite{tversky1974judgment} and that the fine-tuned LLMs (i.e., GPT-3, Mistral-7B \cite{jiang2023mistral}, and T5 \cite{raffel2020exploring}) exhibited more bias compared to their pre-trained counterparts (i.e., GPT-3.5, Mistral-7B-Instruct, Flan-T5, and GPT-4). 

Using a different decision-making context (i.e., operation management (OM)), \citet{chen2023manager} examined 18 common decision-making biases of GPT-3.5 and GPT-4 compared to human managers using classic problems from the literature. They found that GPT-4 displays decision-making patterns that vary by task. In subjective scenarios with uncertain outcomes, GPT shows risk aversion and a preference for certainty. Conversely, when faced with objective tasks, GPT-4 relies on heuristic reasoning. Compared to GPT-3.5, GPT-4  shows both improved accuracy, and increased decision biases, in several contexts. Both GPT models demonstrate a surprising consistency in decision-making across various contexts, including both standard tests, which are commonly reported in the literature and thus already present in the LLMs' training data, and new OM-specific tests that were developed specifically for this study, which were novel and previously unseen by the model.

Similarly, \citet{su2023can} investigated whether GPT-4 can solve the newsvendor problem \cite{schweitzer2000decision} without making biased decisions. They found that it exhibits similar biases to human decision-makers, showing a significantly higher demand-chasing tendency, risk-aversion in high-profit margin scenarios, and loss-aversion in low-profit margin scenarios. However, the model also demonstrated a degree of rationality and responsiveness to incentives, even if they diverged from human predictions based on rational game theory. Notably, GPT-4 did not exhibit certain behavioral biases, such as waste aversion, stockout aversion, underestimated opportunity costs, and minimized ex-post inventory errors. Focusing on economic rationality, \citet{chen2023emergence} explored economic rationality \cite{nishimura2017comprehensive} in GPT-3.5 using a set of curated typical budgetary experiments. They found that the model outperformed human subjects in terms of rationality in four decision-making tasks concerning risk, time, social, and food. However, the performance of the model drops significantly when a less standard presentation of prices is used. Using case studies, \citet{suri2023large} investigated whether GPT-3.5 exhibits decision heuristics similar to humans. Their case study demonstrated that GPT-3.5 shows anchoring Heuristic, representativeness and availability heuristic, framing effect, and the endowment effect.

\subsection{Conclusion}
In summary, modern LLMs demonstrate at least thirteen different human-like heuristics and biases: frame effect, certainty effect, overweighting bias, decoy effect, belief effect, risk aversion, preference for certainty in subjective tasks, heuristic reasoning in objective tasks, demand-chasing, loss-aversion, anchoring effect, representativeness and availability effect, and endowment effect. However, they were also found not to exhibit some other common (in humans) biases, such as reflection effect, isolation effect, magnitude perception, waste aversion, stockout aversion, underestimated opportunity cost, and minimized ex-post inventory errors. Instruction-tuned LLMs show improved performance but also heightened biases, suggesting that the process of instruction fine-tuning may induce increased bias. In more complex decision-making scenarios like the newsvendor problem or operation management, GPT-4 exhibits a blend of human-like biases while also demonstrating rational responses aligned with economic incentives. Interestingly, while LLMs outperform humans in standard decision tasks related to rationality, their performance decreases under less conventional conditions, suggesting a dependency on familiar data presentations. 

Beyond heuristics and biases, sequential decision-making has also become a recent popular subject of LLM research. LLM architectures have inherent advantages in handling sequential information. Unfortunately, most studies in this area use artificial datasets and tasks, and lack human performance as a direct reference for comparison \cite{yang2023foundation}. Additionally, experiments in many of these papers are not hypothesis-driven and cannot be compared with the outcomes of traditional, more rigorously designed psychological experiments. For further details, we refer the interested reader to  \citet{yang2023foundation}.

\section{Reasoning Cognitive Patterns}


\begin{table}[!htp]\centering\caption{Reasoning related studies, with specific type, dataset/tasks used, LLMs involved, associated findings, and corresponding citation. The first three rows, the next four rows, and the last row contain studies related to deductive, inductive, and abductive reasoning, respectively. $\greencheck$ denotes biases found, while $\redcross$ denotes absence of bias. $\greenup$ indicates better than or equivalent to human performance, while $\reddown$ indicates worse than human performance.}\label{tab:reasoning}
\begin{tabularx}{\columnwidth}{Xp{3cm}XXX}
\hline
Type & Dataset & LLMs & Findings & Citation \\
\hline
Deductive &Wason selection task &Guanaco, MPT, BLOOM, Falcon & All models$\reddown$ & \citet{seals2023evaluating} \\
Syllogistic reasoning & Curated dataset derived from BAROCO & RoBERTa, BART, GPT-3.5 & All models$\reddown$  &\citet{ando2023evaluating} \\
Intuitive v.s. rigorous & Cognitive Reflection Test and a semantic illusions task &GPT-2, GPT-3, GPT-3.5, GPT-4 & Pre-ChatGPT: intuitive$\greencheck$, ChatGPT: rigorous$\greencheck$ &\citet{hagendorff2023human} \\
\hline
Inductive & Curated category-based induction dataset inspired by \cite{osherson1990category} & GPT-3, GPT-4 &GPT-4$\greenup$, GPT-3$\reddown$ &\citet{han2024inductive} \\
In-context analogical & Curated RAVEN dataset based on Raven’s Progressive Matrices & OPT 125M, 1.3B, and 13B, text-davinci-002 & All models$\greenup$ &\citet{hu2023context} \\
Analogical &Raven’s Progressive Matrices and a visual analogy problem set &GPT-3, GPT-4 & All models$\greenup$ &\citet{webb2023emergent} \\
Reasoning bias & Variants of the Cognitive Reflection Test and the Linda/Bill problem &GPT-3, GPT-3.5, OPT, BLOOM, LLAMA, VICUNA, GPT-4 & All other models: heuristic reasoning$\greencheck$, GPT-4: heuristic reasoning$\redcross$&\citet{palminteri2023studying} \\
\hline
Bias in abstract and logical reasoning & Curated natural language inference, syllogisms task and Wason Selection Task &Chinchilla, PaLM 2-M, PaLM 2-L, Flan-PaLM 2, and GPT-3.5 & Content effects\greencheck &\citet{dasgupta2022language} \\
\hline
\end{tabularx}
\end{table}

\subsection{Reasoning in humans}
Reasoning may be broadly defined as the process of drawing conclusions based on a combination of axiomatic principles and evidence \cite{sternberg2004we, leighton2004nature}. Reasoning allows us to move from what is already known to making new inferences and incorporating them into our knowledge base, as well as to evaluate proposed hypotheses. In a decision-making framework, we interpret reasoning as the ability to make judgments and choices by integrating salient types of information, weighing evidence, considering alternatives, and predicting potential outcomes. Reasoning may be divided into three types: deductive, inductive, and abductive. Each type is useful in different kinds of contexts and hence operates according to different principles.

Deductive reasoning is the process of using logic to draw conclusions from given observations \cite{johnson1999deductive}. For example, consider the classic example: \textit{(1) All humans are mortal; (2) Socrates is a human; therefore, (3) Socrates is mortal.} The conclusion (3) is valid because it is logically derived from the two premises given. As a result, deductive reasoning is the predominant type of reasoning found in areas like mathematical theorem proving, which require logical rigor. 

Inductive reasoning is the process of reasoning from specific facts or observations to reach a likely conclusion (or `theory') that satisfactorily explains the facts and uses it in an attempt to predict future instances \cite{johnson2000thinking}. Unlike deduction, which gives certainty when provided with true premises, inductive reasoning cannot definitively `prove' its conclusion from premises or evidence; any conclusions must instead be interpreted probabilistically (e.g., as dictated by statistical significance analysis). It is more prevalent in empirical science, where researchers gather data through observation or experimentation to formulate theories that explain the observed phenomena. A well-known example is: after observing many white swans in different locations, one might conclude that all swans are white. The conclusion is inferred from the repeated observation of a phenomenon (although a wrong one). This reflects the fundamental principle of inductive reasoning, which can only allow us to draw a conclusion within limited bounds of certainty, and always carries the caveat that further evidence could reveal exceptions. 

Abductive reasoning aims to recover the `best', usually interpreted as the most plausible, explanation given a set of observations \cite{walton2014abductive}. For example, if a doctor observes a patient with symptoms such as fever, cough, and fatigue, they can use abductive reasoning to conclude that the patient may have the flu, as this would seem to be the the most plausible explanation for the observed symptoms. While lacking the same degree of formalism and traditional philosophical inquiry as inductive and deductive reasoning, it is widely applicable in everyday settings, including humans' ability to infer \textit{plausible causes} by generalizing from sparse data. 

Although each of these types of reasoning can be further subdivided, and the three types do not constitute a strict categorization of reasoning in humans by any means (e.g., other types of reasoning, like deontic reasoning and common sense reasoning, have also been widely studied \cite{beller2008deontic, gordon2017formal}), most studies on LLMs tend to involve one or the other. 

\subsection{Evaluating reasoning in LLMs}
Table \ref{tab:reasoning} summarizes the performance of LLMs, compared with humans, on reasoning tasks. On deductive reasoning, \citet{seals2023evaluating} evaluated the competence of various LLMs, including Guanaco \cite{dettmers2024qlora}, MPT \cite{MosaicML2023Introducing}, BLOOM \cite{le2023bloom}, and Falcon \cite{almazrouei2023falcon}, using the Wason selection task \cite{wason1968reasoning}. They found that when the task is presented in its conventional form, the LLMs show limited performance. When they changed the presentation format, the LLMs did not show significant improvement and exhibited unique reasoning biases that differed from humans.  \citet{ando2023evaluating} evaluated three LLMs (RoBERTa \cite{liu2019roberta}, BART \cite{lewis2019bart}, and GPT-3.5) with a focus on human-like biases (i.e., belief biases, conversion errors, and atmosphere effects) in syllogistic reasoning \cite{van1986essays} using a dataset derived from BAROCO \cite{shikishima2009g}. They found that these models struggled with such problems, and errors caused by various human-like biases heavily influenced their performance. 

In contrast, \citet{hagendorff2023human} explored the differences between intuitive (System 1) and deliberative (System 2) reasoning \cite{kahneman2011thinking} in LLMs using the Cognitive Reflection Test \cite{frederick2005cognitive} and a semantic illusions task \cite{erickson1981words}. They observed that as LLMs get larger and their task comprehension improves, they respond more intuitively, resembling System 1 processing. However, with GPT-3.5 and GPT-4, there was a significant shift towards more deliberative, System 2 like processing. This shift enabled the models to better avoid semantic traps and perform well in cognitive tasks, even without relying explicitly on LLM-specific modalities like chain-of-thought reasoning \cite{wei2022chain}, although such reasoning often appeared in their responses.

On inductive reasoning, \citet{han2024inductive} examined the competence of GPT-3 and GPT-4 using a curated category-based induction task inspired by \cite{osherson1990category}. Their research showed that while GPT-3 faces significant challenges in this area, performing poorly overall and reasoning in a qualitatively different way from humans, GPT-4 demonstrates considerable improvement, achieving performance comparable to that of humans. Additionally, \citet{hu2023context} examined in-context analogical reasoning ability \cite{holyoak1984analogical} in LLMs (i.e., OPT 125M, 1.3B, and 13B \cite{zhang2022opt}, text-davinci-002 \cite{radford2019language}) using the RAVEN dataset \cite{zhang2019raven}, based on Raven’s Progressive Matrices \cite{raven1938raven}. They found that, by encoding perceptual features of problems into language, these LLMs exhibited remarkable zero-shot relational reasoning abilities, often surpassing human performance and approaching the levels of supervised vision-based methods. Furthermore, \citet{webb2023emergent} investigated the analogical reasoning capabilities of GPT-3 and GPT-4 using two zero-shot analogy tasks, including a variant of the Raven’s Progressive Matrices and a visual analogy problem set commonly viewed as one of the best measures of fluid intelligence \cite{snow1984topography}. They found that GPT-3 could match or even surpass human performance in several text-based analogy problems. An initial investigation of GPT-4 revealed its stronger performance on these tasks, suggesting that further scaling of LLMs will likely improve their sensitivity to causal relationships. 

 \citet{palminteri2023studying} studied reasoning biases \cite{simon2013administrative} in LLMs through variants of the Cognitive Reflection Test \cite{branas2019cognitive} and the Linda/Bill problem \cite{tversky1974judgment}, which is known to elicit the conjunction fallacy \cite{sides2002reality}. They observed that earlier LLMs, such as GPT-3 and GPT-3.5, along with open-source models like OPT, BLOOM, LLAMA \cite{touvron2023llama}, and VICUNA \cite{vicuna2023}, demonstrated bounded or heuristic reasoning, relying heavily on intuitive responses and typically under-performing relative to humans. In contrast, more advanced models like GPT-4 exhibited super-human performance on certain tasks, relying minimally on intuitive reasoning. 

Finally, on abductive reasoning, \citet{dasgupta2022language} found that LLMs, including Chinchilla \cite{hoffmann2022training}, PaLM 2-M, PaLM 2-L \cite{chowdhery2023palm}, Flan-PaLM 2 \cite{chung2024scaling}, and GPT-3.5, demonstrate human-like content effects on three logical reasoning tasks that combine the Wason selection task with two other tasks. Specifically, these LLMs, like humans, perform inconsistently across various scenarios and are heavily influenced by the context and content of the stimuli. They face greater challenges with abstract situations or those that contradict their prior understanding of the world, as reflected in their training data.

\subsection{Conclusion}
LLMs exhibit different performance patterns depending on the type of reasoning involved. On deductive reasoning, LLMs struggle with tasks such as the Wason selection task and syllogistic reasoning, often displaying reasoning biases that are different from those shown by humans. However, the more advanced LLMs like GPT-3.5 and GPT-4 show some improvements, particularly in engaging in more deliberative reasoning processes that are similar to humans' system 2 processing. On inductive reasoning, earlier LLMs like GPT-3 faced significant challenges, especially on property induction, but newer models like GPT-4 shows performance almost at par with humans. On analogical reasoning, most LLMs demonstrate strong capabilities, sometimes even surpassing human performance. However, reasoning biases remain a challenge, with LLMs often giving heuristic responses. GPT-4, however, shows signs of surpassing these limitations, delivering super-human performance. Lastly, on abductive reasoning, LLMs demonstrate human-like content effects, and are found to struggle with abstract or counter-intuitive scenarios.

\section{Creativity Cognitive Patterns}


\begin{table}[!htp]\centering\caption{Creativity related studies, with specific type, dataset/tasks used, LLMs involved, associated findings, and corresponding citation. The first three rows, the next two rows, and the last two rows contain studies related to the novel use of objects, creative writing, and assistive creativity, respectively. $\greenup$ indicates better than or equivalent to human performance, while $\reddown$ indicates worse than human performance.}\label{tab:creativity}
\begin{tabularx}{\columnwidth}{Xp{3cm}Xp{2.5cm}X}
\hline
Type & Dataset & LLMs & Findings & Citation \\
\hline
Divergent thinking & Alternative Uses Test & GPT-3 & Originality$\reddown$, surprise$\reddown$, semantic distance$\reddown$, utility$\greenup$&\citet{stevenson2022putting}\\
General & Alternative Uses Test, Torrance Test of Creative Writing, Divergent Associations Task & GPT-4 & Originality$\greenup$, elaboration$\greenup$ &\citet{hubert2023artificial} \\
Novel problem-solving & Novel tool selection task & GPT-3.5, text-davinci-003 & All models$\reddown$ &\citet{yiu2023imitation} \\
\hline
Creative writing & Torrance Test of Creative Writing & GPT-3.5, GPT-4, Claude-v1.3 &All models$\reddown$ &\citet{chakrabarty2023art} \\
Creative writing & Five-sentence creative story task & GPT-3, GPT-4 & All models$\greenup$ & \citet{orwig2024language}\\
\hline
Creative assistance & Five curated tasks involving LLM as creative assistant & GPT-3.5 & Creativity$\greenup$ & \citet{lee2024empirical} \\
Idea generation & Curated dataset & GPT-4 & Novelty$\reddown$, strategic viability$\greenup$, financial value$\greenup$ & \citet{boussioux2024crowdless} \\
\hline
\end{tabularx}
\end{table}

\subsection{Human creativity}

\citet{guilford1950creativity} describes a \textit{creative pattern} as one that  ``is manifest in creative behavior, which includes such activities as inventing, designing, contriving, composing, and planning. People who exhibit these types of behavior to a marked degree are recognized as being creative.'' Creativity, in short, is the ability to produce something that is both original and worthwhile \cite{kozbelt2010theories, sternberg1996investing}. Outcomes of creative pursuits include inventions, discoveries, and artwork. Within decision-making frameworks, creativity is touted as the ability to devise novel alternatives and innovative solutions to problems. One of the ways in which it is measured in humans is divergent production \cite{guilford1950creativity} (more commonly known as `divergent thinking' today), which is the generation of diverse responses when presented with a stimulus. Two prevalent tasks are the Alternative Uses Test (AUT) \cite{guilford1978alternate} and the Torrance Test of Creative Thinking (TTCT) \cite{torrance1966torrance}.

In AUT, participants are tasked with the generation of a diverse array of potential uses for commonplace objects, such as bricks or paper clips. The responses are often evaluated using criteria like originality (the uniqueness of ideas), utility (the practicality of ideas), and surprise (the unpredictability of the ideas). The TTCT comprises two parts: verbal and figural. The verbal component solicits ideas, hypotheses, or solutions from participants using a picture as the stimulus. The responses are assessed on fluency, flexibility, and originality. The figural component asks the participant to complete partially completed shapes or figures. This component evaluates participants' fluency, originality, elaboration, abstractness of titles, resistance to premature closure abilities, and the checklists of creativity strengths. 

\subsection{Creativity in LLMs}
Because of their wide application and success in evaluating human creativity, these two tests have also been employed to measure the creative capabilities of LLMs. \citet{stevenson2022putting} applied AUT to assess GPT-3’s performance and found that human participants scored higher on both originality and surprise, as well as semantic distance, compared to GPT-3. However, GPT-3 received higher utility ratings. Additionally, a negative association between originality and utility was observed in both human and model responses. 

Similarly, \citet{hubert2023artificial} used the AUT task to compare GPT-4’s creativity with that of humans and found that it generated more diverse responses and displayed more elaboration than the human counterpart, exhibiting higher originality when using specific prompts. Additionally, \citet{yiu2023imitation} proposed a novel task that required accomplishing a goal without the typical tool. Their findings revealed that while LLMs, including GPT-3.5 and text-davinci-003, can nearly match humans in recognizing superficial similarities between objects, they significantly lag behind both adults and children when they are asked to choose an unfamiliar tool to solve a problem and often default to conventional solutions rather than novel choices.

Focusing on creative writing, \citet{hubert2023artificial} used the Consequences Task in TTCT and the Divergent Associations Task \cite{olson2021naming} to compare GPT-4’s creativity with that of humans. They revealed that GPT-4 demonstrated greater originality and elaboration than humans across these tasks, even when fluency of responses was controlled for. Extending to TTCT, \citet{chakrabarty2023art} proposed the Torrance Test of Creative Writing (TTCW) to evaluate the creative writing abilities of three LLMs (i.e., GPT-3.5, GPT-4, and Claude-v1.3). Their study found that stories generated by these LLMs were significantly less likely to pass individual TTCW tests compared to those written by human experts. Additionally, \citet{orwig2024language} evaluated two LLMs' (GPT-3 and GPT-4) ability to write creative short stories using the five-sentence creative story task, where the participants are given a three-word prompt and asked to include all three words when writing a short story in approximately five sentences. They find that both LLMs can generate stories that are comparable in creativity to those produced by humans. Interestingly, they also found that GPT-4 was notably consistent in aligning its creativity ratings with those of human evaluators. 

Another interesting area of research is to explore how LLMs can assist humans in creative contexts. \citet{lee2024empirical} evaluated GPT-3.5's ability to assist humans in accomplishing five creative tasks, including choosing a creative gift for a teenager, making a toy, re-purposing unused items, designing an innovative dining table, and re-purposing emotionally significant items. They compared the creativity of ideas generated by humans with the assistance of ChatGPT to those generated using conventional web searches or no technology at all. Their findings demonstrated that ChatGPT significantly improved the creativity of ideas, particularly when it comes to generating incrementally new ideas. In a more nuanced setting, \citet{boussioux2024crowdless} explored human-AI collaboration in generating sustainable, circular economy business ideas using GPT-4. They showed that while human solutions tend to be more novel, human-AI-created solutions often outperformed in terms of strategic viability, environmental and financial value, and overall quality. Additionally, human-AI solutions created through differentiated search, where human-guided prompts progressively directed the LLM to produce outputs increasingly different from the previous versions, outperformed those generated through independent search.

\subsection{Conclusion}
While LLMs like GPT-3 and GPT-3.5 exhibit utility in responses, they often fall short in originality and novelty compared to human creativity in tasks requiring innovative problem-solving. However, more advanced LLMs, such as GPT-4, do tend to give more original and novel solutions than humans. On the other hand, findings on LLMs’ creative writing ability present a more nuanced picture and highlight the need for further investigation. Promisingly, LLMs have shown great potential in collaborative contexts by enhancing human creativity through generation of incrementally new ideas and competence on domain-specific problems and metrics.

\section{Discussion}

Based on these findings, there is promising evidence in favor of LLMs exhibiting all three processes of decision-making, reasoning, and creativity, as emergent patterns. However, the studies also highlight some key limitations. On decision-making, LLMs demonstrate 13 different humanlike heuristics and biases, including frame effect, risk aversion, and anchoring effect. This emergence is likely attributed to the extensive human corpus on which they are trained, which reflects (either explicitly or implicitly) human biases. One reason to be mindful of these biases is that they can lead to undesirable outcomes when deploying them in real-world applications, warranting stronger testing, fail-safes, and ethical reviews. Intriguingly, the \textit{absence} of certain biases (such as \textit{waste aversion} \cite{bolton2012less}) may also yield valuable insights, as they offer insightful commentary, not just of the LLMs' training and underlying neural modeling, but also of the human corpus itself. More practically, a better understanding of how some of these biases are less present in the LLMs than we would expect from human population experiments may help researchers develop pre-training and fine-tuning protocols to proactively exclude these biases from the models if so desired.

The evaluation of reasoning presents a more nuanced picture, mainly because reasoning can be defined in broad ways and involves a wide variety of modalities and cognitive processes, even in humans. On classic deductive reasoning tasks, earlier LLMs struggled, but the performance of the latest models has improved drastically, including on tasks known to be cognitively intensive or even challenging for humans. This shift toward more rigorous, System 2-like reasoning \cite{kahneman2011thinking} in GPT-4 signals the trend of LLMs to mimic more deliberative reasoning. This could potentially be due to the increased use of methodologies like Chain-of-Thought (CoT) being deployed in the prompting of these models. Inductive reasoning tasks appear to be easier for LLMs, achieving even super-human performance on analogical reasoning problems. Analogical reasoning relies heavily on identifying structural similarities between different concepts, and it is possible that LLMs naturally internalize these similarities by learning statistical associations from massive training data. Nevertheless, intuitive and incorrect reasoning processes are still observed in LLMs. This finding, coupled with the phenomenon of hallucinations, which is when LLMs generate outputs that are factually incorrect but still appear very confident in their responses, suggests caution when trusting their reasoning, especially in scenarios that are somewhat novel or implausible.

Commonsense reasoning is yet another challenging aspect of reasoning for LLMs. Unlike deductive or analogical reasoning, commonsense reasoning requires LLMs to use everyday knowledge to make plausible inferences in various scenarios. Current studies on LLMs' commonsense reasoning abilities mainly focus on evaluating their performances on artificial benchmarks, such as the CommonSenseQA \cite{talmor2018commonsenseqa} and the SocialIQa \cite{sap2019socialiqa}. To truly assess the commonsense reasoning abilities of LLMs and compare their performance with humans, researchers should look into tests that are inspired by classic human experiments, as recent pieces have argued \cite{NMICSR,NCSR}. 

The studies that researched creativity collectively reveal a dichotomy: on the one hand, LLMs lack originality and novelty in \textit{divergent} creativity tasks, where they are usually asked to come up with novel use cases of familiar objects. However, in creative \textit{writing}, especially on tasks designed to be relatively open-ended, LLMs like GPT-4 show enormous promise and, in many cases, can be prompted to produce stories that match human creativity \cite{orwig2024language}. This dichotomy can (at least partially) be explained by a language bias: LLMs seem to excel in language-related creative tasks, whereas those that are either multi-modal or abstract may require more (or different) advances. The lack of physical grounding, or embodied cognition, in LLMs has already been noted as an important limitation \cite{carta2023grounding}. Lack of such cognition places obvious constraints on LLMs' ability to create innovative solutions for divergent thinking tasks that require a deep and real-world (including physical) understanding of the objects given. 

Surprisingly, however, LLMs seem to excel in \textit{collaborative} creativity. When used as creative assistants, they have been shown to supplement human effort by generating incrementally novel ideas and providing diverse perspectives. In practice, this makes them prime candidates as \textit{augmented AI}, in jobs ranging from report writing \cite{yeh2024ghostwriter} to software engineering \cite{fan2023large}, both of which are mainstay enterprise applications. It also suggests an important research avenue in both organizational psychology and human-computer interaction, namely, to explore how LLMs can be most effectively integrated into creative industrial processes to foster innovations more rapidly.

Taken together, the studies reveal some important gaps in the current research landscape. Cognitive phenomena like attention \cite{DeWeerd2003-DEWANB}, memory \cite{tulving2000memory}, and representation of knowledge \cite{bruner1956austin}, have not been as extensively studied as the cognitive processes that we reviewed. These phenomena are foundational elements of human cognition and clearly play a role in the three processes that we did cover. For this reason, studying them as individual emergent phenomena in LLMs may have been more difficult. Nevertheless, we maintain that, with the appropriate design of tests, it should be possible to study these phenomena reductively in LLMs. For example, a single promising study focusing on memory \cite{gong2024working} employed the n-back task \cite{kirchner1958age} to evaluate GPT-3.5's capacity for working memory. In another study, \citet{jones2022distrubutional} evaluate knowledge of affordances in GPT-3. Given these possibilities and the currently limited exploration of these phenomena in LLM research, they offer promising avenues for future research. 

Finally, we found that the majority of experiments reported in the literature we reviewed tend to rely heavily on the GPT family of LLMs (which include GPT-3, GPT-3.5, and GPT-4). Although this is probably because of their easy access, other open-source LLMs have not been as extensively tested, which could be a source of bias and non-replication down the line, because GPT models are not (at the time of writing) open in any way. Even the training data used for these models is not publicly documented. Recently, some authors argued that, for scientific reasons, there should be a stronger justification from researchers if they choose to eschew open models \cite{palmer2024using}. More open LLMs offer unique advantages because they are more transparent, offer greater flexibility for customization, and allow researchers to conduct (and replicate) more controlled experiments. At the same time, the GPT family cannot be ignored, owing to its widespread usage in the real world. Fortunately, the two are not in conflict. Given the relative ease of accessing and prompting both open and closed models at the present moment, and the rising number of papers in the AI community that are now choosing to use at least 10+ models in reporting experimental results \cite{gsm8ksymbolic, wei2022chain, mirzadeh2024gsm, rasheed2024large}, we advocate this as a recommended practice in the still-young community of computational psychology.

\bibliography{sn-bibliography}


\begin{thebibliography}{119}
\ifx \bisbn   \undefined \def \bisbn  #1{ISBN #1}\fi
\ifx \binits  \undefined \def \binits#1{#1}\fi
\ifx \bauthor  \undefined \def \bauthor#1{#1}\fi
\ifx \batitle  \undefined \def \batitle#1{#1}\fi
\ifx \bjtitle  \undefined \def \bjtitle#1{#1}\fi
\ifx \bvolume  \undefined \def \bvolume#1{\textbf{#1}}\fi
\ifx \byear  \undefined \def \byear#1{#1}\fi
\ifx \bissue  \undefined \def \bissue#1{#1}\fi
\ifx \bfpage  \undefined \def \bfpage#1{#1}\fi
\ifx \blpage  \undefined \def \blpage #1{#1}\fi
\ifx \burl  \undefined \def \burl#1{\textsf{#1}}\fi
\ifx \doiurl  \undefined \def \doiurl#1{\url{https://doi.org/#1}}\fi
\ifx \betal  \undefined \def \betal{\textit{et al.}}\fi
\ifx \binstitute  \undefined \def \binstitute#1{#1}\fi
\ifx \binstitutionaled  \undefined \def \binstitutionaled#1{#1}\fi
\ifx \bctitle  \undefined \def \bctitle#1{#1}\fi
\ifx \beditor  \undefined \def \beditor#1{#1}\fi
\ifx \bpublisher  \undefined \def \bpublisher#1{#1}\fi
\ifx \bbtitle  \undefined \def \bbtitle#1{#1}\fi
\ifx \bedition  \undefined \def \bedition#1{#1}\fi
\ifx \bseriesno  \undefined \def \bseriesno#1{#1}\fi
\ifx \blocation  \undefined \def \blocation#1{#1}\fi
\ifx \bsertitle  \undefined \def \bsertitle#1{#1}\fi
\ifx \bsnm \undefined \def \bsnm#1{#1}\fi
\ifx \bsuffix \undefined \def \bsuffix#1{#1}\fi
\ifx \bparticle \undefined \def \bparticle#1{#1}\fi
\ifx \barticle \undefined \def \barticle#1{#1}\fi
\bibcommenthead
\ifx \bconfdate \undefined \def \bconfdate #1{#1}\fi
\ifx \botherref \undefined \def \botherref #1{#1}\fi
\ifx \url \undefined \def \url#1{\textsf{#1}}\fi
\ifx \bchapter \undefined \def \bchapter#1{#1}\fi
\ifx \bbook \undefined \def \bbook#1{#1}\fi
\ifx \bcomment \undefined \def \bcomment#1{#1}\fi
\ifx \oauthor \undefined \def \oauthor#1{#1}\fi
\ifx \citeauthoryear \undefined \def \citeauthoryear#1{#1}\fi
\ifx \endbibitem  \undefined \def \endbibitem {}\fi
\ifx \bconflocation  \undefined \def \bconflocation#1{#1}\fi
\ifx \arxivurl  \undefined \def \arxivurl#1{\textsf{#1}}\fi
\csname PreBibitemsHook\endcsname

\bibitem[\protect\citeauthoryear{Brown et~al.}{2020}]{brown2020language}
\begin{barticle}
\bauthor{\bsnm{Brown}, \binits{T.}},
\bauthor{\bsnm{Mann}, \binits{B.}},
\bauthor{\bsnm{Ryder}, \binits{N.}},
\bauthor{\bsnm{Subbiah}, \binits{M.}},
\bauthor{\bsnm{Kaplan}, \binits{J.D.}},
\bauthor{\bsnm{Dhariwal}, \binits{P.}},
\bauthor{\bsnm{Neelakantan}, \binits{A.}},
\bauthor{\bsnm{Shyam}, \binits{P.}},
\bauthor{\bsnm{Sastry}, \binits{G.}},
\bauthor{\bsnm{Askell}, \binits{A.}}, \betal:
\batitle{Language models are few-shot learners}.
\bjtitle{Advances in neural information processing systems}
\bvolume{33},
\bfpage{1877}--\blpage{1901}
(\byear{2020})
\end{barticle}
\endbibitem

\bibitem[\protect\citeauthoryear{Achiam et~al.}{2023}]{achiam2023gpt}
\begin{botherref}
\oauthor{\bsnm{Achiam}, \binits{J.}},
\oauthor{\bsnm{Adler}, \binits{S.}},
\oauthor{\bsnm{Agarwal}, \binits{S.}},
\oauthor{\bsnm{Ahmad}, \binits{L.}},
\oauthor{\bsnm{Akkaya}, \binits{I.}},
\oauthor{\bsnm{Aleman}, \binits{F.L.}},
\oauthor{\bsnm{Almeida}, \binits{D.}},
\oauthor{\bsnm{Altenschmidt}, \binits{J.}},
\oauthor{\bsnm{Altman}, \binits{S.}},
\oauthor{\bsnm{Anadkat}, \binits{S.}}, et al.:
Gpt-4 technical report.
arXiv preprint arXiv:2303.08774
(2023)
\end{botherref}
\endbibitem

\bibitem[\protect\citeauthoryear{Vaswani}{2017}]{vaswani2017attention}
\begin{botherref}
\oauthor{\bsnm{Vaswani}, \binits{A.}}:
Attention is all you need.
Advances in Neural Information Processing Systems
(2017)
\end{botherref}
\endbibitem

\bibitem[\protect\citeauthoryear{Popel et~al.}{2020}]{popel2020transforming}
\begin{barticle}
\bauthor{\bsnm{Popel}, \binits{M.}},
\bauthor{\bsnm{Tomkova}, \binits{M.}},
\bauthor{\bsnm{Tomek}, \binits{J.}},
\bauthor{\bsnm{Kaiser}, \binits{{\L}.}},
\bauthor{\bsnm{Uszkoreit}, \binits{J.}},
\bauthor{\bsnm{Bojar}, \binits{O.}},
\bauthor{\bsnm{{\v{Z}}abokrtsk{\`y}}, \binits{Z.}}:
\batitle{Transforming machine translation: a deep learning system reaches news translation quality comparable to human professionals}.
\bjtitle{Nature communications}
\bvolume{11}(\bissue{1}),
\bfpage{1}--\blpage{15}
(\byear{2020})
\end{barticle}
\endbibitem

\bibitem[\protect\citeauthoryear{Bahdanau}{2014}]{bahdanau2014neural}
\begin{botherref}
\oauthor{\bsnm{Bahdanau}, \binits{D.}}:
Neural machine translation by jointly learning to align and translate.
arXiv preprint arXiv:1409.0473
(2014)
\end{botherref}
\endbibitem

\bibitem[\protect\citeauthoryear{Huang et~al.}{2015}]{huang2015bidirectional}
\begin{botherref}
\oauthor{\bsnm{Huang}, \binits{Z.}},
\oauthor{\bsnm{Xu}, \binits{W.}},
\oauthor{\bsnm{Yu}, \binits{K.}}:
Bidirectional lstm-crf models for sequence tagging.
arXiv preprint arXiv:1508.01991
(2015)
\end{botherref}
\endbibitem

\bibitem[\protect\citeauthoryear{Lin et~al.}{2016}]{lin2016neural}
\begin{bchapter}
\bauthor{\bsnm{Lin}, \binits{Y.}},
\bauthor{\bsnm{Shen}, \binits{S.}},
\bauthor{\bsnm{Liu}, \binits{Z.}},
\bauthor{\bsnm{Luan}, \binits{H.}},
\bauthor{\bsnm{Sun}, \binits{M.}}:
\bctitle{Neural relation extraction with selective attention over instances}.
In: \bbtitle{Proceedings of the 54th Annual Meeting of the Association for Computational Linguistics (Volume 1: Long Papers)},
pp. \bfpage{2124}--\blpage{2133}
(\byear{2016})
\end{bchapter}
\endbibitem

\bibitem[\protect\citeauthoryear{Park et~al.}{2023}]{park2023generative}
\begin{bchapter}
\bauthor{\bsnm{Park}, \binits{J.S.}},
\bauthor{\bsnm{O'Brien}, \binits{J.}},
\bauthor{\bsnm{Cai}, \binits{C.J.}},
\bauthor{\bsnm{Morris}, \binits{M.R.}},
\bauthor{\bsnm{Liang}, \binits{P.}},
\bauthor{\bsnm{Bernstein}, \binits{M.S.}}:
\bctitle{Generative agents: Interactive simulacra of human behavior}.
In: \bbtitle{Proceedings of the 36th Annual Acm Symposium on User Interface Software and Technology},
pp. \bfpage{1}--\blpage{22}
(\byear{2023})
\end{bchapter}
\endbibitem

\bibitem[\protect\citeauthoryear{Strachan et~al.}{2024}]{strachan2024testing}
\begin{botherref}
\oauthor{\bsnm{Strachan}, \binits{J.W.}},
\oauthor{\bsnm{Albergo}, \binits{D.}},
\oauthor{\bsnm{Borghini}, \binits{G.}},
\oauthor{\bsnm{Pansardi}, \binits{O.}},
\oauthor{\bsnm{Scaliti}, \binits{E.}},
\oauthor{\bsnm{Gupta}, \binits{S.}},
\oauthor{\bsnm{Saxena}, \binits{K.}},
\oauthor{\bsnm{Rufo}, \binits{A.}},
\oauthor{\bsnm{Panzeri}, \binits{S.}},
\oauthor{\bsnm{Manzi}, \binits{G.}}, et al.:
Testing theory of mind in large language models and humans.
Nature Human Behaviour,
1--11
(2024)
\end{botherref}
\endbibitem

\bibitem[\protect\citeauthoryear{Bubeck et~al.}{2023}]{bubeck2023sparks}
\begin{botherref}
\oauthor{\bsnm{Bubeck}, \binits{S.}},
\oauthor{\bsnm{Chandrasekaran}, \binits{V.}},
\oauthor{\bsnm{Eldan}, \binits{R.}},
\oauthor{\bsnm{Gehrke}, \binits{J.}},
\oauthor{\bsnm{Horvitz}, \binits{E.}},
\oauthor{\bsnm{Kamar}, \binits{E.}},
\oauthor{\bsnm{Lee}, \binits{P.}},
\oauthor{\bsnm{Lee}, \binits{Y.T.}},
\oauthor{\bsnm{Li}, \binits{Y.}},
\oauthor{\bsnm{Lundberg}, \binits{S.}}, et al.:
Sparks of artificial general intelligence: Early experiments with gpt-4.
arXiv preprint arXiv:2303.12712
(2023)
\end{botherref}
\endbibitem

\bibitem[\protect\citeauthoryear{Sarrion}{2023}]{sarrion2023using}
\begin{bchapter}
\bauthor{\bsnm{Sarrion}, \binits{E.}}:
\bctitle{Using chatgpt for artistic content creation}.
In: \bbtitle{Exploring the Power of ChatGPT: Applications, Techniques, and Implications},
pp. \bfpage{155}--\blpage{170}.
\bpublisher{Springer}, \blocation{???}
(\byear{2023})
\end{bchapter}
\endbibitem

\bibitem[\protect\citeauthoryear{Wei et~al.}{2022}]{wei2022emergent}
\begin{botherref}
\oauthor{\bsnm{Wei}, \binits{J.}},
\oauthor{\bsnm{Tay}, \binits{Y.}},
\oauthor{\bsnm{Bommasani}, \binits{R.}},
\oauthor{\bsnm{Raffel}, \binits{C.}},
\oauthor{\bsnm{Zoph}, \binits{B.}},
\oauthor{\bsnm{Borgeaud}, \binits{S.}},
\oauthor{\bsnm{Yogatama}, \binits{D.}},
\oauthor{\bsnm{Bosma}, \binits{M.}},
\oauthor{\bsnm{Zhou}, \binits{D.}},
\oauthor{\bsnm{Metzler}, \binits{D.}}, et al.:
Emergent abilities of large language models.
arXiv preprint arXiv:2206.07682
(2022)
\end{botherref}
\endbibitem

\bibitem[\protect\citeauthoryear{Schaeffer et~al.}{2024}]{schaeffer2024emergent}
\begin{botherref}
\oauthor{\bsnm{Schaeffer}, \binits{R.}},
\oauthor{\bsnm{Miranda}, \binits{B.}},
\oauthor{\bsnm{Koyejo}, \binits{S.}}:
Are emergent abilities of large language models a mirage?
Advances in Neural Information Processing Systems
\textbf{36}
(2024)
\end{botherref}
\endbibitem

\bibitem[\protect\citeauthoryear{Mikolov}{2013}]{mikolov2013efficient}
\begin{botherref}
\oauthor{\bsnm{Mikolov}, \binits{T.}}:
Efficient estimation of word representations in vector space.
arXiv preprint arXiv:1301.3781
\textbf{3781}
(2013)
\end{botherref}
\endbibitem

\bibitem[\protect\citeauthoryear{Devlin}{2018}]{devlin2018bert}
\begin{botherref}
\oauthor{\bsnm{Devlin}, \binits{J.}}:
Bert: Pre-training of deep bidirectional transformers for language understanding.
arXiv preprint arXiv:1810.04805
(2018)
\end{botherref}
\endbibitem

\bibitem[\protect\citeauthoryear{Rogers et~al.}{2021}]{rogers2021primer}
\begin{barticle}
\bauthor{\bsnm{Rogers}, \binits{A.}},
\bauthor{\bsnm{Kovaleva}, \binits{O.}},
\bauthor{\bsnm{Rumshisky}, \binits{A.}}:
\batitle{A primer in bertology: What we know about how bert works}.
\bjtitle{Transactions of the Association for Computational Linguistics}
\bvolume{8},
\bfpage{842}--\blpage{866}
(\byear{2021})
\end{barticle}
\endbibitem

\bibitem[\protect\citeauthoryear{Shah and Oppenheimer}{2008}]{shah2008heuristics}
\begin{barticle}
\bauthor{\bsnm{Shah}, \binits{A.K.}},
\bauthor{\bsnm{Oppenheimer}, \binits{D.M.}}:
\batitle{Heuristics made easy: an effort-reduction framework.}
\bjtitle{Psychological bulletin}
\bvolume{134}(\bissue{2}),
\bfpage{207}
(\byear{2008})
\end{barticle}
\endbibitem

\bibitem[\protect\citeauthoryear{Kahneman}{1979}]{kahneman1979prospect}
\begin{barticle}
\bauthor{\bsnm{Kahneman}, \binits{D.}}:
\batitle{Prospect theory: An analysis of decisions under risk}.
\bjtitle{Econometrica}
\bvolume{47},
\bfpage{278}
(\byear{1979})
\end{barticle}
\endbibitem

\bibitem[\protect\citeauthoryear{Baron}{2023}]{baron2023thinking}
\begin{bbook}
\bauthor{\bsnm{Baron}, \binits{J.}}:
\bbtitle{Thinking and Deciding}.
\bpublisher{Cambridge University Press}, \blocation{???}
(\byear{2023})
\end{bbook}
\endbibitem

\bibitem[\protect\citeauthoryear{Amabile}{2018}]{amabile2018creativity}
\begin{bbook}
\bauthor{\bsnm{Amabile}, \binits{T.M.}}:
\bbtitle{Creativity in Context: Update to the Social Psychology of Creativity}.
\bpublisher{Routledge}, \blocation{???}
(\byear{2018})
\end{bbook}
\endbibitem

\bibitem[\protect\citeauthoryear{Kahneman and Tversky}{1972}]{kahneman1972subjective}
\begin{barticle}
\bauthor{\bsnm{Kahneman}, \binits{D.}},
\bauthor{\bsnm{Tversky}, \binits{A.}}:
\batitle{Subjective probability: A judgment of representativeness}.
\bjtitle{Cognitive psychology}
\bvolume{3}(\bissue{3}),
\bfpage{430}--\blpage{454}
(\byear{1972})
\end{barticle}
\endbibitem

\bibitem[\protect\citeauthoryear{Binz and Schulz}{2023}]{binz2023a}
\begin{barticle}
\bauthor{\bsnm{Binz}, \binits{M.}},
\bauthor{\bsnm{Schulz}, \binits{E.}}:
\batitle{Using cognitive psychology to understand gpt-3}.
\bjtitle{Proceedings of the National Academy of Sciences}
\bvolume{120}(\bissue{6}),
\bfpage{2218523120}
(\byear{2023})
\end{barticle}
\endbibitem

\bibitem[\protect\citeauthoryear{Itzhak et~al.}{2023}]{itzhak2023instructed}
\begin{botherref}
\oauthor{\bsnm{Itzhak}, \binits{I.}},
\oauthor{\bsnm{Stanovsky}, \binits{G.}},
\oauthor{\bsnm{Rosenfeld}, \binits{N.}},
\oauthor{\bsnm{Belinkov}, \binits{Y.}}:
Instructed to bias: Instruction-tuned language models exhibit emergent cognitive bias.
arXiv preprint arXiv:2308.00225
(2023)
\end{botherref}
\endbibitem

\bibitem[\protect\citeauthoryear{Chen et~al.}{2023}]{chen2023manager}
\begin{botherref}
\oauthor{\bsnm{Chen}, \binits{Y.}},
\oauthor{\bsnm{Andiappan}, \binits{M.}},
\oauthor{\bsnm{Jenkin}, \binits{T.}},
\oauthor{\bsnm{Ovchinnikov}, \binits{A.}}:
A manager and an ai walk into a bar: Does chatgpt make biased decisions like we do?
Available at SSRN 4380365
(2023)
\end{botherref}
\endbibitem

\bibitem[\protect\citeauthoryear{Su et~al.}{2023}]{su2023can}
\begin{botherref}
\oauthor{\bsnm{Su}, \binits{J.}},
\oauthor{\bsnm{Lang}, \binits{Y.}},
\oauthor{\bsnm{Chen}, \binits{K.-Y.}}:
Can ai solve newsvendor problem without making biased decisions? a behavioral experimental study.
A Behavioral Experimental Study (September 1, 2023)
(2023)
\end{botherref}
\endbibitem

\bibitem[\protect\citeauthoryear{Chen et~al.}{2023}]{chen2023emergence}
\begin{botherref}
\oauthor{\bsnm{Chen}, \binits{Y.}},
\oauthor{\bsnm{Liu}, \binits{T.X.}},
\oauthor{\bsnm{Shan}, \binits{Y.}},
\oauthor{\bsnm{Zhong}, \binits{S.}}:
The emergence of economic rationality of gpt.
arXiv preprint arXiv:2305.12763
(2023)
\end{botherref}
\endbibitem

\bibitem[\protect\citeauthoryear{Suri et~al.}{2023}]{suri2023large}
\begin{botherref}
\oauthor{\bsnm{Suri}, \binits{G.}},
\oauthor{\bsnm{Slater}, \binits{L.R.}},
\oauthor{\bsnm{Ziaee}, \binits{A.}},
\oauthor{\bsnm{Nguyen}, \binits{M.}}:
Do large language models show decision heuristics similar to humans? a case study using gpt-3.5.
arXiv preprint arXiv:2305.04400
(2023)
\end{botherref}
\endbibitem

\bibitem[\protect\citeauthoryear{Tversky and Kahneman}{1974}]{tversky1974judgment}
\begin{barticle}
\bauthor{\bsnm{Tversky}, \binits{A.}},
\bauthor{\bsnm{Kahneman}, \binits{D.}}:
\batitle{Judgment under uncertainty: Heuristics and biases: Biases in judgments reveal some heuristics of thinking under uncertainty.}
\bjtitle{science}
\bvolume{185}(\bissue{4157}),
\bfpage{1124}--\blpage{1131}
(\byear{1974})
\end{barticle}
\endbibitem

\bibitem[\protect\citeauthoryear{Jonathan}{2017}]{jonathan2017dual}
\begin{bchapter}
\bauthor{\bsnm{Jonathan}, \binits{B.E.}}:
\bctitle{Dual-process theories}.
In: \bbtitle{International Handbook of Thinking and Reasoning},
pp. \bfpage{151}--\blpage{166}.
\bpublisher{Routledge}, \blocation{???}
(\byear{2017})
\end{bchapter}
\endbibitem

\bibitem[\protect\citeauthoryear{Cialdini and Goldstein}{2004}]{cialdini2004social}
\begin{barticle}
\bauthor{\bsnm{Cialdini}, \binits{R.B.}},
\bauthor{\bsnm{Goldstein}, \binits{N.J.}}:
\batitle{Social influence: Compliance and conformity}.
\bjtitle{Annu. Rev. Psychol.}
\bvolume{55}(\bissue{1}),
\bfpage{591}--\blpage{621}
(\byear{2004})
\end{barticle}
\endbibitem

\bibitem[\protect\citeauthoryear{Glimcher and Rustichini}{2004}]{glimcher2004neuroeconomics}
\begin{barticle}
\bauthor{\bsnm{Glimcher}, \binits{P.W.}},
\bauthor{\bsnm{Rustichini}, \binits{A.}}:
\batitle{Neuroeconomics: the consilience of brain and decision}.
\bjtitle{Science}
\bvolume{306}(\bissue{5695}),
\bfpage{447}--\blpage{452}
(\byear{2004})
\end{barticle}
\endbibitem

\bibitem[\protect\citeauthoryear{Tversky and Kahneman}{1973}]{tversky1973availability}
\begin{barticle}
\bauthor{\bsnm{Tversky}, \binits{A.}},
\bauthor{\bsnm{Kahneman}, \binits{D.}}:
\batitle{Availability: A heuristic for judging frequency and probability}.
\bjtitle{Cognitive psychology}
\bvolume{5}(\bissue{2}),
\bfpage{207}--\blpage{232}
(\byear{1973})
\end{barticle}
\endbibitem

\bibitem[\protect\citeauthoryear{White}{1984}]{white1984representativeness}
\begin{barticle}
\bauthor{\bsnm{White}, \binits{P.A.}}:
\batitle{The representativeness heuristic and the study of judgment under uncertainty.}
\bjtitle{NEW ZEALAND JOURNAL OF PSYCHOLOGY}
\bvolume{13}(\bissue{1}),
\bfpage{1}--\blpage{9}
(\byear{1984})
\end{barticle}
\endbibitem

\bibitem[\protect\citeauthoryear{Furnham and Boo}{2011}]{furnham2011literature}
\begin{barticle}
\bauthor{\bsnm{Furnham}, \binits{A.}},
\bauthor{\bsnm{Boo}, \binits{H.C.}}:
\batitle{A literature review of the anchoring effect}.
\bjtitle{The journal of socio-economics}
\bvolume{40}(\bissue{1}),
\bfpage{35}--\blpage{42}
(\byear{2011})
\end{barticle}
\endbibitem

\bibitem[\protect\citeauthoryear{Gigerenzer and Gaissmaier}{2011}]{gigerenzer2011heuristic}
\begin{barticle}
\bauthor{\bsnm{Gigerenzer}, \binits{G.}},
\bauthor{\bsnm{Gaissmaier}, \binits{W.}}:
\batitle{Heuristic decision making}.
\bjtitle{Annual review of psychology}
\bvolume{62}(\bissue{1}),
\bfpage{451}--\blpage{482}
(\byear{2011})
\end{barticle}
\endbibitem

\bibitem[\protect\citeauthoryear{Tversky and Kahneman}{1983}]{tversky1983extensional}
\begin{barticle}
\bauthor{\bsnm{Tversky}, \binits{A.}},
\bauthor{\bsnm{Kahneman}, \binits{D.}}:
\batitle{Extensional versus intuitive reasoning: The conjunction fallacy in probability judgment.}
\bjtitle{Psychological review}
\bvolume{90}(\bissue{4}),
\bfpage{293}
(\byear{1983})
\end{barticle}
\endbibitem

\bibitem[\protect\citeauthoryear{Tversky and Kahneman}{2015}]{tversky2015causal}
\begin{bchapter}
\bauthor{\bsnm{Tversky}, \binits{A.}},
\bauthor{\bsnm{Kahneman}, \binits{D.}}:
\bctitle{Causal schemas in judgments under uncertainty}.
In: \bbtitle{Progress in Social Psychology},
pp. \bfpage{49}--\blpage{72}.
\bpublisher{Psychology Press}, \blocation{???}
(\byear{2015})
\end{bchapter}
\endbibitem

\bibitem[\protect\citeauthoryear{Peterson et~al.}{2021}]{peterson2021using}
\begin{barticle}
\bauthor{\bsnm{Peterson}, \binits{J.C.}},
\bauthor{\bsnm{Bourgin}, \binits{D.D.}},
\bauthor{\bsnm{Agrawal}, \binits{M.}},
\bauthor{\bsnm{Reichman}, \binits{D.}},
\bauthor{\bsnm{Griffiths}, \binits{T.L.}}:
\batitle{Using large-scale experiments and machine learning to discover theories of human decision-making}.
\bjtitle{Science}
\bvolume{372}(\bissue{6547}),
\bfpage{1209}--\blpage{1214}
(\byear{2021})
\end{barticle}
\endbibitem

\bibitem[\protect\citeauthoryear{Huber et~al.}{1982}]{huber1982adding}
\begin{barticle}
\bauthor{\bsnm{Huber}, \binits{J.}},
\bauthor{\bsnm{Payne}, \binits{J.W.}},
\bauthor{\bsnm{Puto}, \binits{C.}}:
\batitle{Adding asymmetrically dominated alternatives: Violations of regularity and the similarity hypothesis}.
\bjtitle{Journal of consumer research}
\bvolume{9}(\bissue{1}),
\bfpage{90}--\blpage{98}
(\byear{1982})
\end{barticle}
\endbibitem

\bibitem[\protect\citeauthoryear{Evans et~al.}{1983}]{evans1983conflict}
\begin{barticle}
\bauthor{\bsnm{Evans}, \binits{J.S.B.}},
\bauthor{\bsnm{Barston}, \binits{J.L.}},
\bauthor{\bsnm{Pollard}, \binits{P.}}:
\batitle{On the conflict between logic and belief in syllogistic reasoning}.
\bjtitle{Memory \& cognition}
\bvolume{11}(\bissue{3}),
\bfpage{295}--\blpage{306}
(\byear{1983})
\end{barticle}
\endbibitem

\bibitem[\protect\citeauthoryear{Jiang et~al.}{2023}]{jiang2023mistral}
\begin{botherref}
\oauthor{\bsnm{Jiang}, \binits{A.Q.}},
\oauthor{\bsnm{Sablayrolles}, \binits{A.}},
\oauthor{\bsnm{Mensch}, \binits{A.}},
\oauthor{\bsnm{Bamford}, \binits{C.}},
\oauthor{\bsnm{Chaplot}, \binits{D.S.}},
\oauthor{\bsnm{Casas}, \binits{D.d.l.}},
\oauthor{\bsnm{Bressand}, \binits{F.}},
\oauthor{\bsnm{Lengyel}, \binits{G.}},
\oauthor{\bsnm{Lample}, \binits{G.}},
\oauthor{\bsnm{Saulnier}, \binits{L.}}, et al.:
Mistral 7b.
arXiv preprint arXiv:2310.06825
(2023)
\end{botherref}
\endbibitem

\bibitem[\protect\citeauthoryear{Raffel et~al.}{2020}]{raffel2020exploring}
\begin{barticle}
\bauthor{\bsnm{Raffel}, \binits{C.}},
\bauthor{\bsnm{Shazeer}, \binits{N.}},
\bauthor{\bsnm{Roberts}, \binits{A.}},
\bauthor{\bsnm{Lee}, \binits{K.}},
\bauthor{\bsnm{Narang}, \binits{S.}},
\bauthor{\bsnm{Matena}, \binits{M.}},
\bauthor{\bsnm{Zhou}, \binits{Y.}},
\bauthor{\bsnm{Li}, \binits{W.}},
\bauthor{\bsnm{Liu}, \binits{P.J.}}:
\batitle{Exploring the limits of transfer learning with a unified text-to-text transformer}.
\bjtitle{Journal of machine learning research}
\bvolume{21}(\bissue{140}),
\bfpage{1}--\blpage{67}
(\byear{2020})
\end{barticle}
\endbibitem

\bibitem[\protect\citeauthoryear{Schweitzer and Cachon}{2000}]{schweitzer2000decision}
\begin{barticle}
\bauthor{\bsnm{Schweitzer}, \binits{M.E.}},
\bauthor{\bsnm{Cachon}, \binits{G.P.}}:
\batitle{Decision bias in the newsvendor problem with a known demand distribution: Experimental evidence}.
\bjtitle{Management science}
\bvolume{46}(\bissue{3}),
\bfpage{404}--\blpage{420}
(\byear{2000})
\end{barticle}
\endbibitem

\bibitem[\protect\citeauthoryear{Nishimura et~al.}{2017}]{nishimura2017comprehensive}
\begin{barticle}
\bauthor{\bsnm{Nishimura}, \binits{H.}},
\bauthor{\bsnm{Ok}, \binits{E.A.}},
\bauthor{\bsnm{Quah}, \binits{J.K.-H.}}:
\batitle{A comprehensive approach to revealed preference theory}.
\bjtitle{American Economic Review}
\bvolume{107}(\bissue{4}),
\bfpage{1239}--\blpage{1263}
(\byear{2017})
\end{barticle}
\endbibitem

\bibitem[\protect\citeauthoryear{Yang et~al.}{2023}]{yang2023foundation}
\begin{botherref}
\oauthor{\bsnm{Yang}, \binits{S.}},
\oauthor{\bsnm{Nachum}, \binits{O.}},
\oauthor{\bsnm{Du}, \binits{Y.}},
\oauthor{\bsnm{Wei}, \binits{J.}},
\oauthor{\bsnm{Abbeel}, \binits{P.}},
\oauthor{\bsnm{Schuurmans}, \binits{D.}}:
Foundation models for decision making: Problems, methods, and opportunities.
arXiv preprint arXiv:2303.04129
(2023)
\end{botherref}
\endbibitem

\bibitem[\protect\citeauthoryear{Seals and Shalin}{2023}]{seals2023evaluating}
\begin{botherref}
\oauthor{\bsnm{Seals}, \binits{S.}},
\oauthor{\bsnm{Shalin}, \binits{V.L.}}:
Evaluating the deductive competence of large language models.
arXiv preprint arXiv:2309.05452
(2023)
\end{botherref}
\endbibitem

\bibitem[\protect\citeauthoryear{Ando et~al.}{2023}]{ando2023evaluating}
\begin{botherref}
\oauthor{\bsnm{Ando}, \binits{R.}},
\oauthor{\bsnm{Morishita}, \binits{T.}},
\oauthor{\bsnm{Abe}, \binits{H.}},
\oauthor{\bsnm{Mineshima}, \binits{K.}},
\oauthor{\bsnm{Okada}, \binits{M.}}:
Evaluating large language models with neubaroco: Syllogistic reasoning ability and human-like biases.
arXiv preprint arXiv:2306.12567
(2023)
\end{botherref}
\endbibitem

\bibitem[\protect\citeauthoryear{Hagendorff et~al.}{2023}]{hagendorff2023human}
\begin{botherref}
\oauthor{\bsnm{Hagendorff}, \binits{T.}},
\oauthor{\bsnm{Fabi}, \binits{S.}},
\oauthor{\bsnm{Kosinski}, \binits{M.}}:
Human-like intuitive behavior and reasoning biases emerged in large language models but disappeared in chatgpt.
Nature Computational Science,
1--6
(2023)
\end{botherref}
\endbibitem

\bibitem[\protect\citeauthoryear{Osherson et~al.}{1990}]{osherson1990category}
\begin{barticle}
\bauthor{\bsnm{Osherson}, \binits{D.N.}},
\bauthor{\bsnm{Smith}, \binits{E.E.}},
\bauthor{\bsnm{Wilkie}, \binits{O.}},
\bauthor{\bsnm{Lopez}, \binits{A.}},
\bauthor{\bsnm{Shafir}, \binits{E.}}:
\batitle{Category-based induction.}
\bjtitle{Psychological review}
\bvolume{97}(\bissue{2}),
\bfpage{185}
(\byear{1990})
\end{barticle}
\endbibitem

\bibitem[\protect\citeauthoryear{Han et~al.}{2024}]{han2024inductive}
\begin{barticle}
\bauthor{\bsnm{Han}, \binits{S.J.}},
\bauthor{\bsnm{Ransom}, \binits{K.J.}},
\bauthor{\bsnm{Perfors}, \binits{A.}},
\bauthor{\bsnm{Kemp}, \binits{C.}}:
\batitle{Inductive reasoning in humans and large language models}.
\bjtitle{Cognitive Systems Research}
\bvolume{83},
\bfpage{101155}
(\byear{2024})
\end{barticle}
\endbibitem

\bibitem[\protect\citeauthoryear{Hu et~al.}{2023}]{hu2023context}
\begin{botherref}
\oauthor{\bsnm{Hu}, \binits{X.}},
\oauthor{\bsnm{Storks}, \binits{S.}},
\oauthor{\bsnm{Lewis}, \binits{R.L.}},
\oauthor{\bsnm{Chai}, \binits{J.}}:
In-context analogical reasoning with pre-trained language models.
arXiv preprint arXiv:2305.17626
(2023)
\end{botherref}
\endbibitem

\bibitem[\protect\citeauthoryear{Webb et~al.}{2023}]{webb2023emergent}
\begin{botherref}
\oauthor{\bsnm{Webb}, \binits{T.}},
\oauthor{\bsnm{Holyoak}, \binits{K.J.}},
\oauthor{\bsnm{Lu}, \binits{H.}}:
Emergent analogical reasoning in large language models.
Nature Human Behaviour,
1--16
(2023)
\end{botherref}
\endbibitem

\bibitem[\protect\citeauthoryear{Palminteri et~al.}{2023}]{palminteri2023studying}
\begin{botherref}
\oauthor{\bsnm{Palminteri}, \binits{S.}},
\oauthor{\bsnm{Yax}, \binits{N.}},
\oauthor{\bsnm{Anllo}, \binits{H.}}:
Studying and improving reasoning in humans and machines
(2023)
\end{botherref}
\endbibitem

\bibitem[\protect\citeauthoryear{Dasgupta et~al.}{2022}]{dasgupta2022language}
\begin{botherref}
\oauthor{\bsnm{Dasgupta}, \binits{I.}},
\oauthor{\bsnm{Lampinen}, \binits{A.K.}},
\oauthor{\bsnm{Chan}, \binits{S.C.}},
\oauthor{\bsnm{Creswell}, \binits{A.}},
\oauthor{\bsnm{Kumaran}, \binits{D.}},
\oauthor{\bsnm{McClelland}, \binits{J.L.}},
\oauthor{\bsnm{Hill}, \binits{F.}}:
Language models show human-like content effects on reasoning.
arXiv preprint arXiv:2207.07051
(2022)
\end{botherref}
\endbibitem

\bibitem[\protect\citeauthoryear{Sternberg}{2004}]{sternberg2004we}
\begin{botherref}
\oauthor{\bsnm{Sternberg}, \binits{R.J.}}:
What do we know about the nature of reasoning?
(2004)
\end{botherref}
\endbibitem

\bibitem[\protect\citeauthoryear{Leighton and Sternberg}{2004}]{leighton2004nature}
\begin{bbook}
\bauthor{\bsnm{Leighton}, \binits{J.P.}},
\bauthor{\bsnm{Sternberg}, \binits{R.J.}}:
\bbtitle{The Nature of Reasoning}.
\bpublisher{Cambridge University Press}, \blocation{???}
(\byear{2004})
\end{bbook}
\endbibitem

\bibitem[\protect\citeauthoryear{Johnson-Laird}{1999}]{johnson1999deductive}
\begin{barticle}
\bauthor{\bsnm{Johnson-Laird}, \binits{P.N.}}:
\batitle{Deductive reasoning}.
\bjtitle{Annual review of psychology}
\bvolume{50}(\bissue{1}),
\bfpage{109}--\blpage{135}
(\byear{1999})
\end{barticle}
\endbibitem

\bibitem[\protect\citeauthoryear{Johnson-Laird}{2000}]{johnson2000thinking}
\begin{bbook}
\bauthor{\bsnm{Johnson-Laird}, \binits{P.N.}}:
\bbtitle{Thinking: Reasoning.}
\bpublisher{American Psychological Association}, \blocation{???}
(\byear{2000})
\end{bbook}
\endbibitem

\bibitem[\protect\citeauthoryear{Walton}{2014}]{walton2014abductive}
\begin{bbook}
\bauthor{\bsnm{Walton}, \binits{D.}}:
\bbtitle{Abductive Reasoning}.
\bpublisher{University of Alabama Press}, \blocation{???}
(\byear{2014})
\end{bbook}
\endbibitem

\bibitem[\protect\citeauthoryear{Beller}{2008}]{beller2008deontic}
\begin{barticle}
\bauthor{\bsnm{Beller}, \binits{S.}}:
\batitle{Deontic norms, deontic reasoning, and deontic conditionals}.
\bjtitle{Thinking \& Reasoning}
\bvolume{14}(\bissue{4}),
\bfpage{305}--\blpage{341}
(\byear{2008})
\end{barticle}
\endbibitem

\bibitem[\protect\citeauthoryear{Gordon and Hobbs}{2017}]{gordon2017formal}
\begin{bbook}
\bauthor{\bsnm{Gordon}, \binits{A.S.}},
\bauthor{\bsnm{Hobbs}, \binits{J.R.}}:
\bbtitle{A Formal Theory of Commonsense Psychology: How People Think People Think}.
\bpublisher{Cambridge University Press}, \blocation{???}
(\byear{2017})
\end{bbook}
\endbibitem

\bibitem[\protect\citeauthoryear{Dettmers et~al.}{2024}]{dettmers2024qlora}
\begin{botherref}
\oauthor{\bsnm{Dettmers}, \binits{T.}},
\oauthor{\bsnm{Pagnoni}, \binits{A.}},
\oauthor{\bsnm{Holtzman}, \binits{A.}},
\oauthor{\bsnm{Zettlemoyer}, \binits{L.}}:
Qlora: Efficient finetuning of quantized llms.
Advances in Neural Information Processing Systems
\textbf{36}
(2024)
\end{botherref}
\endbibitem

\bibitem[\protect\citeauthoryear{Team}{2023}]{MosaicML2023Introducing}
\begin{botherref}
\oauthor{\bsnm{Team}, \binits{M.N.}}:
Introducing MPT-7B: A New Standard for Open-Source, Commercially Usable LLMs.
Accessed: 2023-05-05.
\url{www.mosaicml.com/blog/mpt-7b}
Accessed 2023-05-05
\end{botherref}
\endbibitem

\bibitem[\protect\citeauthoryear{Le~Scao et~al.}{2023}]{le2023bloom}
\begin{botherref}
\oauthor{\bsnm{Le~Scao}, \binits{T.}},
\oauthor{\bsnm{Fan}, \binits{A.}},
\oauthor{\bsnm{Akiki}, \binits{C.}},
\oauthor{\bsnm{Pavlick}, \binits{E.}},
\oauthor{\bsnm{Ili{\'c}}, \binits{S.}},
\oauthor{\bsnm{Hesslow}, \binits{D.}},
\oauthor{\bsnm{Castagn{\'e}}, \binits{R.}},
\oauthor{\bsnm{Luccioni}, \binits{A.S.}},
\oauthor{\bsnm{Yvon}, \binits{F.}},
\oauthor{\bsnm{Gall{\'e}}, \binits{M.}}, et al.:
Bloom: A 176b-parameter open-access multilingual language model
(2023)
\end{botherref}
\endbibitem

\bibitem[\protect\citeauthoryear{Almazrouei et~al.}{2023}]{almazrouei2023falcon}
\begin{botherref}
\oauthor{\bsnm{Almazrouei}, \binits{E.}},
\oauthor{\bsnm{Alobeidli}, \binits{H.}},
\oauthor{\bsnm{Alshamsi}, \binits{A.}},
\oauthor{\bsnm{Cappelli}, \binits{A.}},
\oauthor{\bsnm{Cojocaru}, \binits{R.}},
\oauthor{\bsnm{Debbah}, \binits{M.}},
\oauthor{\bsnm{Goffinet}, \binits{{\'E}.}},
\oauthor{\bsnm{Hesslow}, \binits{D.}},
\oauthor{\bsnm{Launay}, \binits{J.}},
\oauthor{\bsnm{Malartic}, \binits{Q.}}, et al.:
The falcon series of open language models.
arXiv preprint arXiv:2311.16867
(2023)
\end{botherref}
\endbibitem

\bibitem[\protect\citeauthoryear{Wason}{1968}]{wason1968reasoning}
\begin{barticle}
\bauthor{\bsnm{Wason}, \binits{P.C.}}:
\batitle{Reasoning about a rule}.
\bjtitle{Quarterly journal of experimental psychology}
\bvolume{20}(\bissue{3}),
\bfpage{273}--\blpage{281}
(\byear{1968})
\end{barticle}
\endbibitem

\bibitem[\protect\citeauthoryear{Liu}{2019}]{liu2019roberta}
\begin{botherref}
\oauthor{\bsnm{Liu}, \binits{Y.}}:
Roberta: A robustly optimized bert pretraining approach.
arXiv preprint arXiv:1907.11692
(2019)
\end{botherref}
\endbibitem

\bibitem[\protect\citeauthoryear{Lewis}{2019}]{lewis2019bart}
\begin{botherref}
\oauthor{\bsnm{Lewis}, \binits{M.}}:
Bart: Denoising sequence-to-sequence pre-training for natural language generation, translation, and comprehension.
arXiv preprint arXiv:1910.13461
(2019)
\end{botherref}
\endbibitem

\bibitem[\protect\citeauthoryear{Van~Benthem et~al.}{1986}]{van1986essays}
\begin{bbook}
\bauthor{\bsnm{Van~Benthem}, \binits{J.}}, \betal:
\bbtitle{Essays in Logical Semantics}
vol. \bseriesno{29}.
\bpublisher{Springer}, \blocation{???}
(\byear{1986})
\end{bbook}
\endbibitem

\bibitem[\protect\citeauthoryear{Shikishima et~al.}{2009}]{shikishima2009g}
\begin{barticle}
\bauthor{\bsnm{Shikishima}, \binits{C.}},
\bauthor{\bsnm{Hiraishi}, \binits{K.}},
\bauthor{\bsnm{Yamagata}, \binits{S.}},
\bauthor{\bsnm{Sugimoto}, \binits{Y.}},
\bauthor{\bsnm{Takemura}, \binits{R.}},
\bauthor{\bsnm{Ozaki}, \binits{K.}},
\bauthor{\bsnm{Okada}, \binits{M.}},
\bauthor{\bsnm{Toda}, \binits{T.}},
\bauthor{\bsnm{Ando}, \binits{J.}}:
\batitle{Is g an entity? a japanese twin study using syllogisms and intelligence tests}.
\bjtitle{Intelligence}
\bvolume{37}(\bissue{3}),
\bfpage{256}--\blpage{267}
(\byear{2009})
\end{barticle}
\endbibitem

\bibitem[\protect\citeauthoryear{Kahneman}{2011}]{kahneman2011thinking}
\begin{bbook}
\bauthor{\bsnm{Kahneman}, \binits{D.}}:
\bbtitle{Thinking, Fast and Slow}.
\bpublisher{macmillan}, \blocation{???}
(\byear{2011})
\end{bbook}
\endbibitem

\bibitem[\protect\citeauthoryear{Frederick}{2005}]{frederick2005cognitive}
\begin{barticle}
\bauthor{\bsnm{Frederick}, \binits{S.}}:
\batitle{Cognitive reflection and decision making}.
\bjtitle{Journal of Economic perspectives}
\bvolume{19}(\bissue{4}),
\bfpage{25}--\blpage{42}
(\byear{2005})
\end{barticle}
\endbibitem

\bibitem[\protect\citeauthoryear{Erickson and Mattson}{1981}]{erickson1981words}
\begin{barticle}
\bauthor{\bsnm{Erickson}, \binits{T.D.}},
\bauthor{\bsnm{Mattson}, \binits{M.E.}}:
\batitle{From words to meaning: A semantic illusion}.
\bjtitle{Journal of Verbal Learning and Verbal Behavior}
\bvolume{20}(\bissue{5}),
\bfpage{540}--\blpage{551}
(\byear{1981})
\end{barticle}
\endbibitem

\bibitem[\protect\citeauthoryear{Wei et~al.}{2022}]{wei2022chain}
\begin{barticle}
\bauthor{\bsnm{Wei}, \binits{J.}},
\bauthor{\bsnm{Wang}, \binits{X.}},
\bauthor{\bsnm{Schuurmans}, \binits{D.}},
\bauthor{\bsnm{Bosma}, \binits{M.}},
\bauthor{\bsnm{Xia}, \binits{F.}},
\bauthor{\bsnm{Chi}, \binits{E.}},
\bauthor{\bsnm{Le}, \binits{Q.V.}},
\bauthor{\bsnm{Zhou}, \binits{D.}}, \betal:
\batitle{Chain-of-thought prompting elicits reasoning in large language models}.
\bjtitle{Advances in neural information processing systems}
\bvolume{35},
\bfpage{24824}--\blpage{24837}
(\byear{2022})
\end{barticle}
\endbibitem

\bibitem[\protect\citeauthoryear{Holyoak}{1984}]{holyoak1984analogical}
\begin{barticle}
\bauthor{\bsnm{Holyoak}, \binits{K.J.}}:
\batitle{Analogical thinking and human intelligence}.
\bjtitle{Advances in the psychology of human intelligence}
\bvolume{2},
\bfpage{199}--\blpage{230}
(\byear{1984})
\end{barticle}
\endbibitem

\bibitem[\protect\citeauthoryear{Zhang et~al.}{2022}]{zhang2022opt}
\begin{botherref}
\oauthor{\bsnm{Zhang}, \binits{S.}},
\oauthor{\bsnm{Roller}, \binits{S.}},
\oauthor{\bsnm{Goyal}, \binits{N.}},
\oauthor{\bsnm{Artetxe}, \binits{M.}},
\oauthor{\bsnm{Chen}, \binits{M.}},
\oauthor{\bsnm{Chen}, \binits{S.}},
\oauthor{\bsnm{Dewan}, \binits{C.}},
\oauthor{\bsnm{Diab}, \binits{M.}},
\oauthor{\bsnm{Li}, \binits{X.}},
\oauthor{\bsnm{Lin}, \binits{X.V.}}, et al.:
Opt: Open pre-trained transformer language models.
arXiv preprint arXiv:2205.01068
(2022)
\end{botherref}
\endbibitem

\bibitem[\protect\citeauthoryear{Radford et~al.}{2019}]{radford2019language}
\begin{barticle}
\bauthor{\bsnm{Radford}, \binits{A.}},
\bauthor{\bsnm{Wu}, \binits{J.}},
\bauthor{\bsnm{Child}, \binits{R.}},
\bauthor{\bsnm{Luan}, \binits{D.}},
\bauthor{\bsnm{Amodei}, \binits{D.}},
\bauthor{\bsnm{Sutskever}, \binits{I.}}, \betal:
\batitle{Language models are unsupervised multitask learners}.
\bjtitle{OpenAI blog}
\bvolume{1}(\bissue{8}),
\bfpage{9}
(\byear{2019})
\end{barticle}
\endbibitem

\bibitem[\protect\citeauthoryear{Zhang et~al.}{2019}]{zhang2019raven}
\begin{bchapter}
\bauthor{\bsnm{Zhang}, \binits{C.}},
\bauthor{\bsnm{Gao}, \binits{F.}},
\bauthor{\bsnm{Jia}, \binits{B.}},
\bauthor{\bsnm{Zhu}, \binits{Y.}},
\bauthor{\bsnm{Zhu}, \binits{S.-C.}}:
\bctitle{Raven: A dataset for relational and analogical visual reasoning}.
In: \bbtitle{Proceedings of the IEEE/CVF Conference on Computer Vision and Pattern Recognition},
pp. \bfpage{5317}--\blpage{5327}
(\byear{2019})
\end{bchapter}
\endbibitem

\bibitem[\protect\citeauthoryear{Raven}{1938}]{raven1938raven}
\begin{botherref}
\oauthor{\bsnm{Raven}, \binits{J.}}:
Raven's progressive matrices: Western Psychological Services Los Angeles.
CA
(1938)
\end{botherref}
\endbibitem

\bibitem[\protect\citeauthoryear{Snow et~al.}{1984}]{snow1984topography}
\begin{barticle}
\bauthor{\bsnm{Snow}, \binits{R.E.}},
\bauthor{\bsnm{Kyllonen}, \binits{P.C.}},
\bauthor{\bsnm{Marshalek}, \binits{B.}}, \betal:
\batitle{The topography of ability and learning correlations}.
\bjtitle{Advances in the psychology of human intelligence}
\bvolume{2}(\bissue{S 47}),
\bfpage{103}
(\byear{1984})
\end{barticle}
\endbibitem

\bibitem[\protect\citeauthoryear{Simon}{2013}]{simon2013administrative}
\begin{bbook}
\bauthor{\bsnm{Simon}, \binits{H.A.}}:
\bbtitle{Administrative Behavior}.
\bpublisher{Simon and Schuster}, \blocation{???}
(\byear{2013})
\end{bbook}
\endbibitem

\bibitem[\protect\citeauthoryear{Bra{\~n}as-Garza et~al.}{2019}]{branas2019cognitive}
\begin{barticle}
\bauthor{\bsnm{Bra{\~n}as-Garza}, \binits{P.}},
\bauthor{\bsnm{Kujal}, \binits{P.}},
\bauthor{\bsnm{Lenkei}, \binits{B.}}:
\batitle{Cognitive reflection test: Whom, how, when}.
\bjtitle{Journal of Behavioral and Experimental Economics}
\bvolume{82},
\bfpage{101455}
(\byear{2019})
\end{barticle}
\endbibitem

\bibitem[\protect\citeauthoryear{Sides et~al.}{2002}]{sides2002reality}
\begin{barticle}
\bauthor{\bsnm{Sides}, \binits{A.}},
\bauthor{\bsnm{Osherson}, \binits{D.}},
\bauthor{\bsnm{Bonini}, \binits{N.}},
\bauthor{\bsnm{Viale}, \binits{R.}}:
\batitle{On the reality of the conjunction fallacy}.
\bjtitle{Memory \& cognition}
\bvolume{30},
\bfpage{191}--\blpage{198}
(\byear{2002})
\end{barticle}
\endbibitem

\bibitem[\protect\citeauthoryear{Touvron et~al.}{2023}]{touvron2023llama}
\begin{botherref}
\oauthor{\bsnm{Touvron}, \binits{H.}},
\oauthor{\bsnm{Lavril}, \binits{T.}},
\oauthor{\bsnm{Izacard}, \binits{G.}},
\oauthor{\bsnm{Martinet}, \binits{X.}},
\oauthor{\bsnm{Lachaux}, \binits{M.-A.}},
\oauthor{\bsnm{Lacroix}, \binits{T.}},
\oauthor{\bsnm{Rozi{\`e}re}, \binits{B.}},
\oauthor{\bsnm{Goyal}, \binits{N.}},
\oauthor{\bsnm{Hambro}, \binits{E.}},
\oauthor{\bsnm{Azhar}, \binits{F.}}, et al.:
Llama: Open and efficient foundation language models.
arXiv preprint arXiv:2302.13971
(2023)
\end{botherref}
\endbibitem

\bibitem[\protect\citeauthoryear{Chiang et~al.}{2023}]{vicuna2023}
\begin{botherref}
\oauthor{\bsnm{Chiang}, \binits{W.-L.}},
\oauthor{\bsnm{Li}, \binits{Z.}},
\oauthor{\bsnm{Lin}, \binits{Z.}},
\oauthor{\bsnm{Sheng}, \binits{Y.}},
\oauthor{\bsnm{Wu}, \binits{Z.}},
\oauthor{\bsnm{Zhang}, \binits{H.}},
\oauthor{\bsnm{Zheng}, \binits{L.}},
\oauthor{\bsnm{Zhuang}, \binits{S.}},
\oauthor{\bsnm{Zhuang}, \binits{Y.}},
\oauthor{\bsnm{Gonzalez}, \binits{J.E.}},
\oauthor{\bsnm{Stoica}, \binits{I.}},
\oauthor{\bsnm{Xing}, \binits{E.P.}}:
Vicuna: An Open-Source Chatbot Impressing GPT-4 with 90\%* ChatGPT Quality
(2023).
\url{https://lmsys.org/blog/2023-03-30-vicuna/}
\end{botherref}
\endbibitem

\bibitem[\protect\citeauthoryear{Hoffmann et~al.}{2022}]{hoffmann2022training}
\begin{botherref}
\oauthor{\bsnm{Hoffmann}, \binits{J.}},
\oauthor{\bsnm{Borgeaud}, \binits{S.}},
\oauthor{\bsnm{Mensch}, \binits{A.}},
\oauthor{\bsnm{Buchatskaya}, \binits{E.}},
\oauthor{\bsnm{Cai}, \binits{T.}},
\oauthor{\bsnm{Rutherford}, \binits{E.}},
\oauthor{\bsnm{Casas}, \binits{D.d.L.}},
\oauthor{\bsnm{Hendricks}, \binits{L.A.}},
\oauthor{\bsnm{Welbl}, \binits{J.}},
\oauthor{\bsnm{Clark}, \binits{A.}}, et al.:
Training compute-optimal large language models.
arXiv preprint arXiv:2203.15556
(2022)
\end{botherref}
\endbibitem

\bibitem[\protect\citeauthoryear{Chowdhery et~al.}{2023}]{chowdhery2023palm}
\begin{barticle}
\bauthor{\bsnm{Chowdhery}, \binits{A.}},
\bauthor{\bsnm{Narang}, \binits{S.}},
\bauthor{\bsnm{Devlin}, \binits{J.}},
\bauthor{\bsnm{Bosma}, \binits{M.}},
\bauthor{\bsnm{Mishra}, \binits{G.}},
\bauthor{\bsnm{Roberts}, \binits{A.}},
\bauthor{\bsnm{Barham}, \binits{P.}},
\bauthor{\bsnm{Chung}, \binits{H.W.}},
\bauthor{\bsnm{Sutton}, \binits{C.}},
\bauthor{\bsnm{Gehrmann}, \binits{S.}}, \betal:
\batitle{Palm: Scaling language modeling with pathways}.
\bjtitle{Journal of Machine Learning Research}
\bvolume{24}(\bissue{240}),
\bfpage{1}--\blpage{113}
(\byear{2023})
\end{barticle}
\endbibitem

\bibitem[\protect\citeauthoryear{Chung et~al.}{2024}]{chung2024scaling}
\begin{barticle}
\bauthor{\bsnm{Chung}, \binits{H.W.}},
\bauthor{\bsnm{Hou}, \binits{L.}},
\bauthor{\bsnm{Longpre}, \binits{S.}},
\bauthor{\bsnm{Zoph}, \binits{B.}},
\bauthor{\bsnm{Tay}, \binits{Y.}},
\bauthor{\bsnm{Fedus}, \binits{W.}},
\bauthor{\bsnm{Li}, \binits{Y.}},
\bauthor{\bsnm{Wang}, \binits{X.}},
\bauthor{\bsnm{Dehghani}, \binits{M.}},
\bauthor{\bsnm{Brahma}, \binits{S.}}, \betal:
\batitle{Scaling instruction-finetuned language models}.
\bjtitle{Journal of Machine Learning Research}
\bvolume{25}(\bissue{70}),
\bfpage{1}--\blpage{53}
(\byear{2024})
\end{barticle}
\endbibitem

\bibitem[\protect\citeauthoryear{Stevenson et~al.}{2022}]{stevenson2022putting}
\begin{botherref}
\oauthor{\bsnm{Stevenson}, \binits{C.}},
\oauthor{\bsnm{Smal}, \binits{I.}},
\oauthor{\bsnm{Baas}, \binits{M.}},
\oauthor{\bsnm{Grasman}, \binits{R.}},
\oauthor{\bsnm{Maas}, \binits{H.}}:
Putting gpt-3's creativity to the (alternative uses) test.
arXiv preprint arXiv:2206.08932
(2022)
\end{botherref}
\endbibitem

\bibitem[\protect\citeauthoryear{Hubert et~al.}{2023}]{hubert2023artificial}
\begin{botherref}
\oauthor{\bsnm{Hubert}, \binits{K.}},
\oauthor{\bsnm{Awa}, \binits{K.N.}},
\oauthor{\bsnm{Zabelina}, \binits{D.}}:
Artificial intelligence is more creative than humans: A cognitive science perspective on the current state of generative language models
(2023)
\end{botherref}
\endbibitem

\bibitem[\protect\citeauthoryear{Yiu et~al.}{2023}]{yiu2023imitation}
\begin{botherref}
\oauthor{\bsnm{Yiu}, \binits{E.}},
\oauthor{\bsnm{Kosoy}, \binits{E.}},
\oauthor{\bsnm{Gopnik}, \binits{A.}}:
Imitation versus innovation: What children can do that large language and language-and-vision models cannot (yet)?
arXiv preprint arXiv:2305.07666
(2023)
\end{botherref}
\endbibitem

\bibitem[\protect\citeauthoryear{Chakrabarty et~al.}{2023}]{chakrabarty2023art}
\begin{botherref}
\oauthor{\bsnm{Chakrabarty}, \binits{T.}},
\oauthor{\bsnm{Laban}, \binits{P.}},
\oauthor{\bsnm{Agarwal}, \binits{D.}},
\oauthor{\bsnm{Muresan}, \binits{S.}},
\oauthor{\bsnm{Wu}, \binits{C.-S.}}:
Art or artifice? large language models and the false promise of creativity.
arXiv preprint arXiv:2309.14556
(2023)
\end{botherref}
\endbibitem

\bibitem[\protect\citeauthoryear{Orwig et~al.}{2024}]{orwig2024language}
\begin{barticle}
\bauthor{\bsnm{Orwig}, \binits{W.}},
\bauthor{\bsnm{Edenbaum}, \binits{E.R.}},
\bauthor{\bsnm{Greene}, \binits{J.D.}},
\bauthor{\bsnm{Schacter}, \binits{D.L.}}:
\batitle{The language of creativity: Evidence from humans and large language models}.
\bjtitle{The Journal of creative behavior}
\bvolume{58}(\bissue{1}),
\bfpage{128}--\blpage{136}
(\byear{2024})
\end{barticle}
\endbibitem

\bibitem[\protect\citeauthoryear{Lee and Chung}{2024}]{lee2024empirical}
\begin{botherref}
\oauthor{\bsnm{Lee}, \binits{B.C.}},
\oauthor{\bsnm{Chung}, \binits{J.}}:
An empirical investigation of the impact of chatgpt on creativity.
Nature Human Behaviour,
1--9
(2024)
\end{botherref}
\endbibitem

\bibitem[\protect\citeauthoryear{Boussioux et~al.}{2024}]{boussioux2024crowdless}
\begin{barticle}
\bauthor{\bsnm{Boussioux}, \binits{L.}},
\bauthor{\bsnm{Lane}, \binits{J.N.}},
\bauthor{\bsnm{Zhang}, \binits{M.}},
\bauthor{\bsnm{Jacimovic}, \binits{V.}},
\bauthor{\bsnm{Lakhani}, \binits{K.R.}}:
\batitle{The crowdless future? generative ai and creative problem-solving}.
\bjtitle{Organization Science}
\bvolume{35}(\bissue{5}),
\bfpage{1589}--\blpage{1607}
(\byear{2024})
\end{barticle}
\endbibitem

\bibitem[\protect\citeauthoryear{Guilford}{1950}]{guilford1950creativity}
\begin{botherref}
\oauthor{\bsnm{Guilford}, \binits{J.}}:
Creativity. American Psychology. 5 (9), 444--454
(1950)
\end{botherref}
\endbibitem

\bibitem[\protect\citeauthoryear{Kozbelt et~al.}{2010}]{kozbelt2010theories}
\begin{barticle}
\bauthor{\bsnm{Kozbelt}, \binits{A.}},
\bauthor{\bsnm{Beghetto}, \binits{R.A.}},
\bauthor{\bsnm{Runco}, \binits{M.A.}}:
\batitle{Theories of creativity}.
\bjtitle{The Cambridge handbook of creativity}
\bvolume{2},
\bfpage{20}--\blpage{47}
(\byear{2010})
\end{barticle}
\endbibitem

\bibitem[\protect\citeauthoryear{Sternberg and Lubart}{1996}]{sternberg1996investing}
\begin{barticle}
\bauthor{\bsnm{Sternberg}, \binits{R.J.}},
\bauthor{\bsnm{Lubart}, \binits{T.I.}}:
\batitle{Investing in creativity.}
\bjtitle{American psychologist}
\bvolume{51}(\bissue{7}),
\bfpage{677}
(\byear{1996})
\end{barticle}
\endbibitem

\bibitem[\protect\citeauthoryear{Guilford et~al.}{1978}]{guilford1978alternate}
\begin{botherref}
\oauthor{\bsnm{Guilford}, \binits{J.P.}},
\oauthor{\bsnm{Christensen}, \binits{P.R.}},
\oauthor{\bsnm{Merrifield}, \binits{P.R.}},
\oauthor{\bsnm{Wilson}, \binits{R.C.}}:
Alternate uses
(1978)
\end{botherref}
\endbibitem

\bibitem[\protect\citeauthoryear{Torrance}{1966}]{torrance1966torrance}
\begin{botherref}
\oauthor{\bsnm{Torrance}, \binits{E.P.}}:
Torrance tests of creative thinking.
Educational and Psychological Measurement
(1966)
\end{botherref}
\endbibitem

\bibitem[\protect\citeauthoryear{Olson et~al.}{2021}]{olson2021naming}
\begin{barticle}
\bauthor{\bsnm{Olson}, \binits{J.A.}},
\bauthor{\bsnm{Nahas}, \binits{J.}},
\bauthor{\bsnm{Chmoulevitch}, \binits{D.}},
\bauthor{\bsnm{Cropper}, \binits{S.J.}},
\bauthor{\bsnm{Webb}, \binits{M.E.}}:
\batitle{Naming unrelated words predicts creativity}.
\bjtitle{Proceedings of the National Academy of Sciences}
\bvolume{118}(\bissue{25}),
\bfpage{2022340118}
(\byear{2021})
\end{barticle}
\endbibitem

\bibitem[\protect\citeauthoryear{Bolton and Alba}{2012}]{bolton2012less}
\begin{barticle}
\bauthor{\bsnm{Bolton}, \binits{L.E.}},
\bauthor{\bsnm{Alba}, \binits{J.W.}}:
\batitle{When less is more: Consumer aversion to unused utility}.
\bjtitle{Journal of consumer psychology}
\bvolume{22}(\bissue{3}),
\bfpage{369}--\blpage{383}
(\byear{2012})
\end{barticle}
\endbibitem

\bibitem[\protect\citeauthoryear{Talmor et~al.}{2018}]{talmor2018commonsenseqa}
\begin{botherref}
\oauthor{\bsnm{Talmor}, \binits{A.}},
\oauthor{\bsnm{Herzig}, \binits{J.}},
\oauthor{\bsnm{Lourie}, \binits{N.}},
\oauthor{\bsnm{Berant}, \binits{J.}}:
Commonsenseqa: A question answering challenge targeting commonsense knowledge.
arXiv preprint arXiv:1811.00937
(2018)
\end{botherref}
\endbibitem

\bibitem[\protect\citeauthoryear{Sap et~al.}{2019}]{sap2019socialiqa}
\begin{botherref}
\oauthor{\bsnm{Sap}, \binits{M.}},
\oauthor{\bsnm{Rashkin}, \binits{H.}},
\oauthor{\bsnm{Chen}, \binits{D.}},
\oauthor{\bsnm{LeBras}, \binits{R.}},
\oauthor{\bsnm{Choi}, \binits{Y.}}:
Socialiqa: Commonsense reasoning about social interactions.
arXiv preprint arXiv:1904.09728
(2019)
\end{botherref}
\endbibitem

\bibitem[\protect\citeauthoryear{Kejriwal et~al.}{2022}]{NMICSR}
\begin{barticle}
\bauthor{\bsnm{Kejriwal}, \binits{M.}},
\bauthor{\bsnm{Santos}, \binits{H.}},
\bauthor{\bsnm{Mulvehill}, \binits{A.M.}},
\bauthor{\bsnm{McGuinness}, \binits{D.L.}}:
\batitle{Designing a strong test for measuring true common-sense reasoning}.
\bjtitle{Nature Machine Intelligence}
\bvolume{4}(\bissue{4}),
\bfpage{318}--\blpage{322}
(\byear{2022})
\end{barticle}
\endbibitem

\bibitem[\protect\citeauthoryear{Kejriwal et~al.}{2024}]{NCSR}
\begin{barticle}
\bauthor{\bsnm{Kejriwal}, \binits{M.}},
\bauthor{\bsnm{Santos}, \binits{H.}},
\bauthor{\bsnm{Mulvehill}, \binits{A.M.}},
\bauthor{\bsnm{Shen}, \binits{K.}},
\bauthor{\bsnm{McGuinness}, \binits{D.L.}},
\bauthor{\bsnm{Lieberman}, \binits{H.}}:
\batitle{Can ai have common sense? finding out will be key to achieving machine intelligence}.
\bjtitle{Nature}
\bvolume{634}(\bissue{8033}),
\bfpage{291}--\blpage{294}
(\byear{2024})
\end{barticle}
\endbibitem

\bibitem[\protect\citeauthoryear{Carta et~al.}{2023}]{carta2023grounding}
\begin{bchapter}
\bauthor{\bsnm{Carta}, \binits{T.}},
\bauthor{\bsnm{Romac}, \binits{C.}},
\bauthor{\bsnm{Wolf}, \binits{T.}},
\bauthor{\bsnm{Lamprier}, \binits{S.}},
\bauthor{\bsnm{Sigaud}, \binits{O.}},
\bauthor{\bsnm{Oudeyer}, \binits{P.-Y.}}:
\bctitle{Grounding large language models in interactive environments with online reinforcement learning}.
In: \bbtitle{International Conference on Machine Learning},
pp. \bfpage{3676}--\blpage{3713}
(\byear{2023}).
\bcomment{PMLR}
\end{bchapter}
\endbibitem

\bibitem[\protect\citeauthoryear{Yeh et~al.}{2024}]{yeh2024ghostwriter}
\begin{botherref}
\oauthor{\bsnm{Yeh}, \binits{C.}},
\oauthor{\bsnm{Ramos}, \binits{G.}},
\oauthor{\bsnm{Ng}, \binits{R.}},
\oauthor{\bsnm{Huntington}, \binits{A.}},
\oauthor{\bsnm{Banks}, \binits{R.}}:
Ghostwriter: Augmenting collaborative human-ai writing experiences through personalization and agency.
arXiv preprint arXiv:2402.08855
(2024)
\end{botherref}
\endbibitem

\bibitem[\protect\citeauthoryear{Fan et~al.}{2023}]{fan2023large}
\begin{bchapter}
\bauthor{\bsnm{Fan}, \binits{A.}},
\bauthor{\bsnm{Gokkaya}, \binits{B.}},
\bauthor{\bsnm{Harman}, \binits{M.}},
\bauthor{\bsnm{Lyubarskiy}, \binits{M.}},
\bauthor{\bsnm{Sengupta}, \binits{S.}},
\bauthor{\bsnm{Yoo}, \binits{S.}},
\bauthor{\bsnm{Zhang}, \binits{J.M.}}:
\bctitle{Large language models for software engineering: Survey and open problems}.
In: \bbtitle{2023 IEEE/ACM International Conference on Software Engineering: Future of Software Engineering (ICSE-FoSE)},
pp. \bfpage{31}--\blpage{53}
(\byear{2023}).
\bcomment{IEEE}
\end{bchapter}
\endbibitem

\bibitem[\protect\citeauthoryear{Weerd}{2003}]{DeWeerd2003-DEWANB}
\begin{bchapter}
\bauthor{\bsnm{Weerd}, \binits{P.D.}}:
\bctitle{Attention, neural basis of}.
In: \beditor{\bsnm{Nadel}, \binits{L.}} (ed.)
\bbtitle{Encyclopedia of Cognitive Science},
pp. \bfpage{238}--\blpage{246}.
\bpublisher{Nature Publishing Group}, \blocation{???}
(\byear{2003})
\end{bchapter}
\endbibitem

\bibitem[\protect\citeauthoryear{Tulving}{2000}]{tulving2000memory}
\begin{botherref}
\oauthor{\bsnm{Tulving}, \binits{E.}}:
Memory: An overview.
(2000)
\end{botherref}
\endbibitem

\bibitem[\protect\citeauthoryear{Bruner et~al.}{1956}]{bruner1956austin}
\begin{barticle}
\bauthor{\bsnm{Bruner}, \binits{J.S.}},
\bauthor{\bsnm{Goodnow}, \binits{J.J.}},
\bauthor{\bsnm{George}, \binits{A.}}:
\batitle{Austin. a study of thinking}.
\bjtitle{New York: John Wiley \& Sons, Inc}
\bvolume{14},
\bfpage{330}
(\byear{1956})
\end{barticle}
\endbibitem

\bibitem[\protect\citeauthoryear{Gong et~al.}{2024}]{gong2024working}
\begin{bchapter}
\bauthor{\bsnm{Gong}, \binits{D.}},
\bauthor{\bsnm{Wan}, \binits{X.}},
\bauthor{\bsnm{Wang}, \binits{D.}}:
\bctitle{Working memory capacity of chatgpt: An empirical study}.
In: \bbtitle{Proceedings of the AAAI Conference on Artificial Intelligence},
vol. \bseriesno{38},
pp. \bfpage{10048}--\blpage{10056}
(\byear{2024})
\end{bchapter}
\endbibitem

\bibitem[\protect\citeauthoryear{Kirchner}{1958}]{kirchner1958age}
\begin{barticle}
\bauthor{\bsnm{Kirchner}, \binits{W.K.}}:
\batitle{Age differences in short-term retention of rapidly changing information.}
\bjtitle{Journal of experimental psychology}
\bvolume{55}(\bissue{4}),
\bfpage{352}
(\byear{1958})
\end{barticle}
\endbibitem

\bibitem[\protect\citeauthoryear{Jones et~al.}{2022}]{jones2022distrubutional}
\begin{bchapter}
\bauthor{\bsnm{Jones}, \binits{C.R.}},
\bauthor{\bsnm{Chang}, \binits{T.A.}},
\bauthor{\bsnm{Coulson}, \binits{S.}},
\bauthor{\bsnm{Michaelov}, \binits{J.A.}},
\bauthor{\bsnm{Trott}, \binits{S.}},
\bauthor{\bsnm{Bergen}, \binits{B.}}:
\bctitle{Distrubutional semantics still can't account for affordances}.
In: \bbtitle{Proceedings of the Annual Meeting of the Cognitive Science Society},
vol. \bseriesno{44}
(\byear{2022})
\end{bchapter}
\endbibitem

\bibitem[\protect\citeauthoryear{Palmer et~al.}{2024}]{palmer2024using}
\begin{barticle}
\bauthor{\bsnm{Palmer}, \binits{A.}},
\bauthor{\bsnm{Smith}, \binits{N.A.}},
\bauthor{\bsnm{Spirling}, \binits{A.}}:
\batitle{Using proprietary language models in academic research requires explicit justification}.
\bjtitle{Nature Computational Science}
\bvolume{4}(\bissue{1}),
\bfpage{2}--\blpage{3}
(\byear{2024})
\end{barticle}
\endbibitem

\bibitem[\protect\citeauthoryear{Mirzadeh et~al.}{2024a}]{gsm8ksymbolic}
\begin{botherref}
\oauthor{\bsnm{Mirzadeh}, \binits{I.}},
\oauthor{\bsnm{Alizadeh}, \binits{K.}},
\oauthor{\bsnm{Shahrokhi}, \binits{H.}},
\oauthor{\bsnm{Tuzel}, \binits{O.}},
\oauthor{\bsnm{Bengio}, \binits{S.}},
\oauthor{\bsnm{Farajtabar}, \binits{M.}}:
Gsm-symbolic: Understanding the limitations of mathematical reasoning in large language models.
arXiv preprint arXiv:2410.05229
(2024)
\end{botherref}
\endbibitem

\bibitem[\protect\citeauthoryear{Mirzadeh et~al.}{2024b}]{mirzadeh2024gsm}
\begin{botherref}
\oauthor{\bsnm{Mirzadeh}, \binits{I.}},
\oauthor{\bsnm{Alizadeh}, \binits{K.}},
\oauthor{\bsnm{Shahrokhi}, \binits{H.}},
\oauthor{\bsnm{Tuzel}, \binits{O.}},
\oauthor{\bsnm{Bengio}, \binits{S.}},
\oauthor{\bsnm{Farajtabar}, \binits{M.}}:
Gsm-symbolic: Understanding the limitations of mathematical reasoning in large language models.
arXiv preprint arXiv:2410.05229
(2024)
\end{botherref}
\endbibitem

\bibitem[\protect\citeauthoryear{Rasheed et~al.}{2024}]{rasheed2024large}
\begin{botherref}
\oauthor{\bsnm{Rasheed}, \binits{Z.}},
\oauthor{\bsnm{Waseem}, \binits{M.}},
\oauthor{\bsnm{Syst{\"a}}, \binits{K.}},
\oauthor{\bsnm{Abrahamsson}, \binits{P.}}:
Large language model evaluation via multi ai agents: Preliminary results.
arXiv preprint arXiv:2404.01023
(2024)
\end{botherref}
\endbibitem

\end{thebibliography}

\end{document}